%% file: arxiv.tex
\documentclass[10pt,twocolumn,letterpaper]{article}

\usepackage{cvpr}              

\usepackage{graphicx}
\usepackage{amsmath}
\usepackage{amssymb}
\usepackage{booktabs}
\usepackage[accsupp]{axessibility}
\usepackage{multirow}
\input{math.tex}

\usepackage[pagebackref,breaklinks,colorlinks]{hyperref}

\usepackage[capitalize]{cleveref}
\crefname{section}{Sec.}{Secs.}
\Crefname{section}{Section}{Sections}
\Crefname{table}{Table}{Tables}
\crefname{table}{Tab.}{Tabs.}

\def\cvprPaperID{7243} 
\def\confName{CVPR}
\def\confYear{2023}
\def\arxiv{arxiv}

\begin{document}

\makeatletter
\newcommand{\printfnsymbol}[1]{%
  \textsuperscript{\@fnsymbol{#1}}%
}
\makeatother

\title{Siamese DETR}

\author{
    Zeren Chen\textsuperscript{\rm 1, 2}\thanks{Equal contribution.},
    Gengshi Huang\textsuperscript{\rm 2}\printfnsymbol{1}, 
    Wei Li\textsuperscript{\rm 3}, 
    Jianing Teng\textsuperscript{\rm 2}, 
    Kun Wang\textsuperscript{\rm 2}, \\
    Jing Shao\textsuperscript{\rm 2}, 
    Chen Change Loy\textsuperscript{\rm 3},
    Lu Sheng\textsuperscript{\rm 1}\thanks{Corresponding author.}\\
    \large\textsuperscript{\rm 1} School of Software, Beihang University,
    \large\textsuperscript{\rm 2} SenseTime Research, \\
    \large\textsuperscript{\rm 3} S-Lab, Nanyang Technological University. \\
    \small\texttt{\{czr1604,lsheng\}@buaa.edu.cn},\;\;\texttt{\{wei.l,ccloy\}@ntu.edu.sg},\;\;\texttt{huanggengshi@gmail.com}\\
    \small\texttt{wangkun@sensetime.com},\;\;\texttt{\{tengjianing,shaojing\}@senseauto.com}.
}
\maketitle

\input{sections/0_abstract}

\input{sections/1_intro}
\input{sections/2_related_work}
\input{sections/3_method}
\input{sections/4_exp}
\input{sections/5_conclusion}

\input{sections/6_acknownledgement}
\input{sections/7_appendix}

{\small
\bibliographystyle{ieee_fullname}
\bibliography{egbib}
}

\end{document}

%% file: math.tex
\usepackage{amsmath,amsfonts,bm}

\def\eqref#1{equation~\ref{#1}}

\def\1{\bm{1}}

\def\vb{{\bm{b}}}
\def\vc{{\bm{c}}}

\def\vh{{\bm{h}}}

\def\vk{{\bm{k}}}

\def\vp{{\bm{p}}}
\def\vq{{\bm{q}}}

\def\vv{{\bm{v}}}

\def\vx{{\bm{x}}}

\def\vz{{\bm{z}}}
\def\vphi{{\bm{\phi}}}

\DeclareMathAlphabet{\mathsfit}{\encodingdefault}{\sfdefault}{m}{sl}
\SetMathAlphabet{\mathsfit}{bold}{\encodingdefault}{\sfdefault}{bx}{n}



%% file: sections/0_abstract.tex
\begin{abstract}
Recent self-supervised methods are mainly designed for representation learning with the base model, \emph{e.g.}, ResNets or ViTs. They cannot be easily transferred to DETR, with task-specific Transformer modules.
In this work, we present \textbf{Siamese DETR}, a \textbf{Siamese} self-supervised pretraining approach for the Transformer architecture in \textbf{DETR}. 
We consider learning view-invariant and detection-oriented representations simultaneously through two complementary tasks, \emph{i.e.}, localization and discrimination, in a novel multi-view learning framework.
Two self-supervised pretext tasks are designed:
(i) \textbf{Multi-View Region Detection} aims at learning to localize regions-of-interest between augmented views of the input, and (ii) \textbf{Multi-View Semantic Discrimination} attempts to improve object-level discrimination for each region. 
The proposed Siamese DETR achieves state-of-the-art transfer performance on COCO and PASCAL VOC detection using different DETR variants in all setups. 
Code is available at \href{https://github.com/Zx55/SiameseDETR}{https://github.com/Zx55/SiameseDETR}.
\end{abstract}

%% file: sections/1_intro.tex
\section{Introduction}\label{sec:intro}

Object detection with Transformers (DETR)~\cite{carion2020end} combines convolutional neural networks (CNNs) and Transformer-based encoder-decoders, viewing object detection as an end-to-end set prediction problem. 
Despite its impressive performance, DETR and its variants still rely on large-scale, high-quality training data. It generally requires huge cost and effort to collect such massive well-annotated datasets, which can be prohibited in some privacy-sensitive applications such as medical imaging and video surveillance.

\begin{figure}[!t]
\begin{center}
\includegraphics[width=\linewidth]{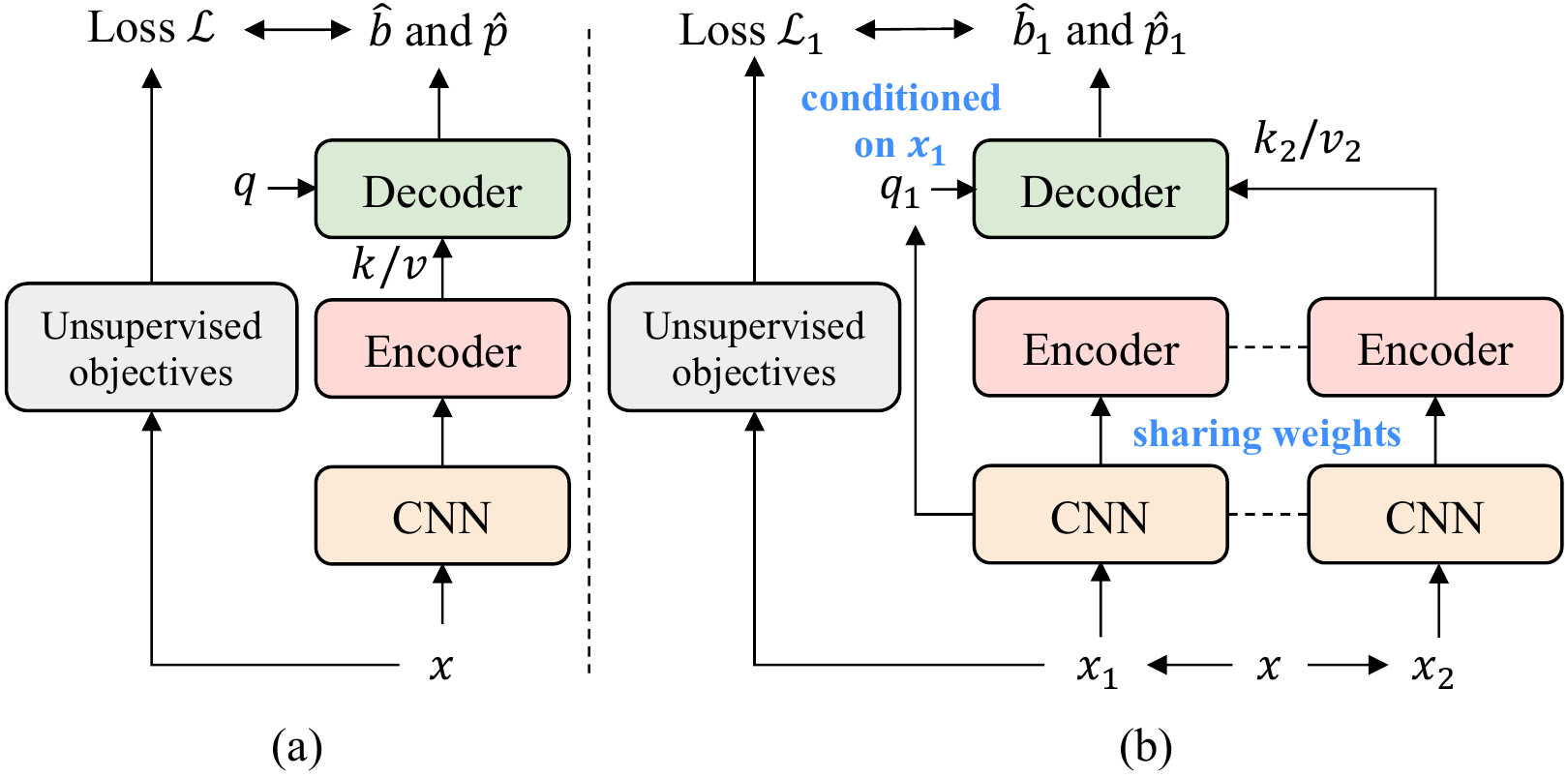}
\vskip -0.2cm
\end{center}
    \caption{
        Comparison between single-view and multi-view detection pretraining for DETR. \textbf{(a)} The single-view framework, \emph{e.g.}, UP-DETR \cite{dai2021up} and DETReg \cite{bar2022detreg}, perform self-supervised representation learning using unsupervised objectives generated on the single view, \emph{e.g.}, random patches (UP-DETR) or pseudo labels (DETReg), leading to a small information gain during pretraining. \textbf{(b)} The proposed multi-view Siamese DETR for DETR pretraining. Here, $\hat{b}$ and $\hat{p}$ denote box and semantic predictions. $\vq$, $\vk$ and $\vv$ denote query, key and value in DETR, respectively.
    }
    \label{fig:siamese_detr_concept}
\end{figure}

Recent progress in multi-view self-supervised representation learning~\cite{chen2020simple, he2020momentum,chen2020improved,caron2020unsupervised,chen2021exploring, li2021prototypical,grill2020bootstrap} can potentially alleviate the appetite for labeled data in training DETR for object detection.
However, these self-supervised learning approaches mainly focus on learning generalizable representations with base models, such as ResNets~\cite{he2016deep} and ViTs \cite{dosovitskiy2020image}.
It is unclear how these approaches can be effectively extended to DETR with task-specific Transformers modules that are tailored for end-to-end object detection. 

Designing self-supervised pretext tasks for pretraining the Transformers in DETR is a challenging and practical problem, demanding representations that could benefit object detection, beyond just learning generic representation. 
Several attempts have been made to address this issue. For example, UP-DETR \cite{dai2021up} introduces an unsupervised pretext task based on random query patch detection, predicting bounding boxes of randomly-cropped query patches in the given image. Recent DETReg \cite{bar2022detreg} employs a pre-trained SwAV \cite{swav} and offline Selective Search proposals \cite{selective_search} to provide pseudo labels for DETR pertaining. In general, both UP-DETR and DETReg follow a single-view pretraining paradigm (see Figure \ref{fig:siamese_detr_concept} \textbf{(a)}), without exploring the ability of learning view-invariant representations demonstrated in existing multi-view self-supervised approaches.

In this work, we are interested in investigating the effectiveness of multi-view self-supervised learning for DETR pre-training.
Different from conventional multi-view framework \cite{he2020momentum, chen2020simple, swav}, we combine the Siamese network with the cross-attention mechanism in DETR, presenting a Siamese self-supervised pretraining approach, named Siamese DETR, with two proposed self-supervised pretext tasks dedicated to view-invariant detection pretraining.
Specifically, given each unlabeled image, we follow \cite{wei2021aligning, bar2022detreg} to obtain the offline object proposals and generate two augmented views guided by Intersection over Union (IoU) thresholds.
As illustrated in Figure \ref{fig:siamese_detr_concept} \textbf{(b)}, by directly locating the query regions between augmented views and maximizing the discriminative information at both global and regional levels, Siamese DETR can learn view-invariant representations with localization and discrimination that are aligned with downstream object detection tasks during pretraining.
Our contributions can be summarized as below:

\begin{itemize} 

    \item We propose a novel Siamese self-supervised approach for the Transformers in DETR, which jointly learns view-invariant representations with discrimination and localization. In particular, we contribute two new designs of self-supervised pretext tasks specialized for multi-view detection pretraining.
    
    \item Without bells and whistles, Siamese DETR outperforms UP-DETR \cite{dai2021up} and DETReg \cite{bar2022detreg} with multiple DETR variants, such as Conditional \cite{meng2021conditional} and Deformable \cite{zhu2020deformable} DETR, on the COCO and PASCAL VOC benchmarks, demonstrating the effectiveness and versatility of our designs.
    
\end{itemize}

%% file: sections/2_related_work.tex
\section{Related Work}

\noindent\textbf{Object Detection with Transformers.} 
DETR \cite{carion2020end} integrates CNNs with Transformers \cite{vaswani2017attention}, effectively eliminating the need for hand-crafted components, such as rule-based training target assignment, anchor generation, and non-maximum suppression.
Several recent DETR variants \cite{zhu2020deformable, meng2021conditional, Gao_2021_ICCV, li2022dn, zhang2022dino} have been proposed to improve the attention mechanism and bipartite matching in Transformers.
For example, \cite{zhu2020deformable} only attend to a small set of key sampling points around a reference for faster convergence. 
\cite{meng2021conditional} learn a conditional spatial query from the decoder embedding for decoder multi-head cross-attention.
\cite{li2022dn} introduce a denoising pipeline to reduce the difficulty of bipartite matching.

In contrast, we explore another paradigm to improve the representations of Transformers for DETR via self-supervised pretraining.
\cite{dai2021up} design a pretext task based on random query patch detection. 
\cite{bar2022detreg} train DETR using pseudo labels generated by pretrained SwAV \cite{swav} and offline proposals \cite{selective_search}.
While similarly following the existing \textit{pre-train and fine-tune} paradigm, our work significantly differs from UP-DETR and DETReg.
To our knowledge, we make the first attempt to combine the Siamese network with a cross-attention mechanism in DETR, enabling the model to learn view-invariant localizing ability during pretraining.

\vspace{+1mm}
\noindent\textbf{Self-supervised Pretraining.} 
One of the main approaches for self-supervised learning is to compare different augmented views of the same data instances in the representation space. 
Some notable studies include that by \cite{chen2020simple}, which presents a simple framework by removing the requirements of specialized architectures or memory banks. 
\cite{he2020momentum} employ a momentum encoder with a dynamic dictionary look-up and retrieve more negative samples by using large dictionary sizes.
\cite{chen2021exploring} explore simple Siamese networks to learn meaningful representations with positive pairs only. 
\cite{caron2020unsupervised} enforce consistency between cluster assignments produced for different augmentations of the same image. 
In addition, several attempts \cite{xie2021detco, zhao2021self, wang2021dense, wei2021aligning, Henaff_2021_ICCV} have been made to learn detection-oriented representations directly using intrinsic cues, such as mask predictions \cite{zhao2021self, Henaff_2021_ICCV}, offline region proposals \cite{wei2021aligning, xie2021unsupervised}, and joint global-local partitions \cite{wang2021dense, xie2021detco}.
While different in the specific learning strategies, all these works focus on learning discriminative representations for base models, which are insufficient for transfer learning in DETR.
In addition, the pretext tasks in these methods cannot be directly applied to the Transformers in DETR.

\vspace{+1mm}
\noindent\textbf{Siamese Networks.}
Siamese networks \cite{bromley1993signature} are weight-sharing neural networks and usually take two inputs for comparison, which are widely adopted in many applications, such as face verification \cite{taigman2014deepface}, person re-identification \cite{zheng2019re}, one-shot learning \cite{koch2015siamese} and semi-supervise learning \cite{liu2021unbiased, xu2021end}.
Recent advances in self-supervised learning \cite{chen2020improved, chen2020simple, chen2021exploring, grill2020bootstrap} are also built upon Siamese networks, motivating us to explore the Siamese architecture for pretraining the Transformers in DETR.

%% file: sections/3_method.tex
\begin{figure*}[t!]
    \centering
    \includegraphics[width=\linewidth]{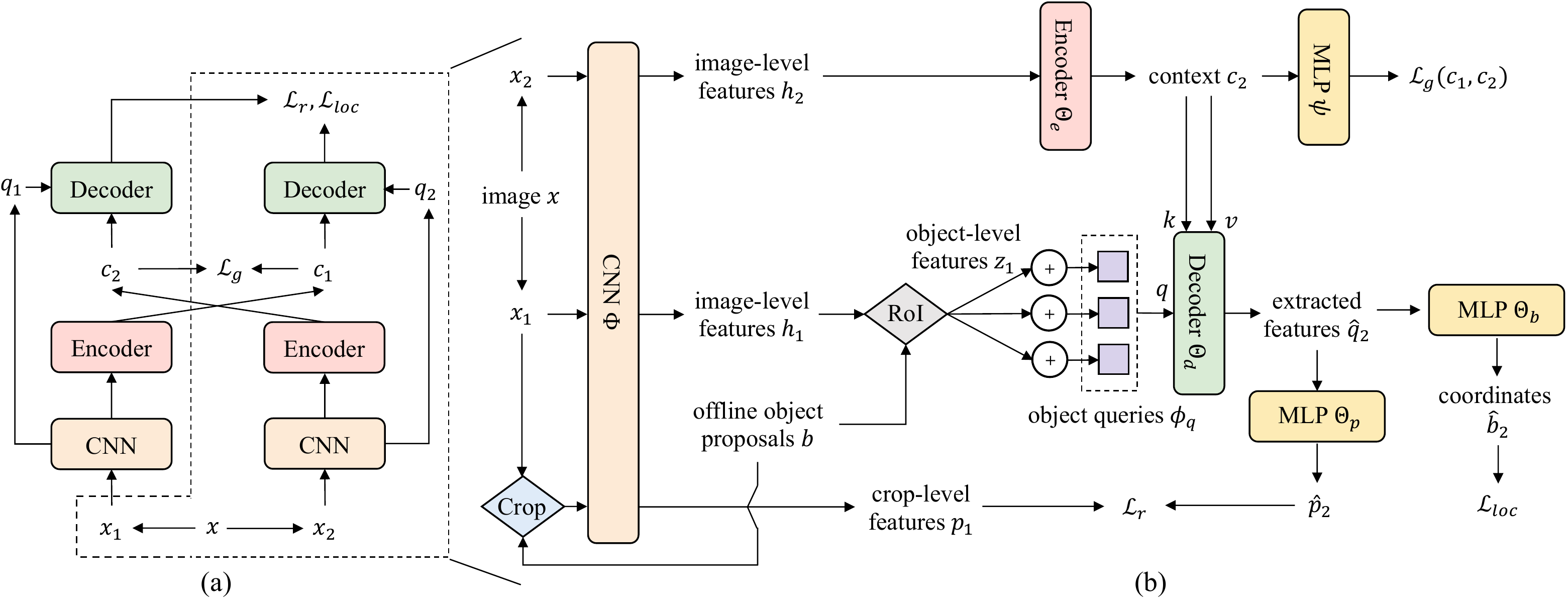}
    \caption{
        \textbf{(a)} Overall architecture of proposed Siamese DETR. 
        \textbf{(b)} The forward process of one view in our symmetrical pipeline.
        We perform region detection ($\mathcal{L}_{loc}$) and semantic discrimination ($\mathcal{L}_g$ and $\mathcal{L}_r$) in a multi-view fashion. 
        Given the conditional input of region features in one view, we aim at locating and discriminating their corresponding regions in another view.
    }
    \label{fig:overview}
\end{figure*}

\section{Siamese DETR}
\label{sec:siamese_detr}
An overview of our Siamese DETR architecture is presented in Figure \ref{fig:overview}, which illustrates the main pipeline of our multi-view detection pretraining.
We first revisit the DETR in Section \ref{sec:ca_detr}. 
We then describe the view construction algorithm in Section \ref{subsec:views} and the multi-view pretraining paradigm of Siamese DETR in Section \ref{sec:mvdp}, powered by two specially designed self-supervised pretext tasks for learning to detect objects.

\subsection{Revisiting DETR}
\label{sec:ca_detr}
A typical DETR model consists of two modules: 
(i) a backbone model, \emph{i.e.}, CNNs, for feature extraction, 
(ii) Transformers with encoder-decoders architecture for set prediction, built by stacking multi-head attentions \cite{vaswani2017attention}. 
The backbone model extracts the image-level features $\vh = \texttt{Backbone}(\vx)$ for a given image $\vx\in\mathbb{R}^{3\times H_0\times W_0}$.
Then, the Transformer encoder takes 
$\vh$ as inputs, encoding the image features as global context $\vc\in\mathbb{R}^{C\times H_1\times W_1}$: 
\begin{equation}
    \vc = \texttt{Encoder}(\vh).
\end{equation}
Note that we omit the positional embedding $\vphi_p$ in the description for clarity.
The cross-attention mechanism is a general form of multi-head attention ($\texttt{MHA}$) in the Transformer decoder, which calculates the weighted sum $\widehat{\vq}\in\mathbb{R}^{N\times C}$ between the flattened global context $\vc\in\mathbb{R}^{H_1W_1\times C}$ and $N$ object queries $\vphi_q\in\mathbb{R}^{N\times C}$ for further box and categorical prediction.
\begin{equation}
\label{eq:ca}
\begin{aligned}
    \widehat{\vq} &= \texttt{CrossAtten}(\vc, \vphi_q) = \texttt{MHA}(\vq, \vk, \vv) \\
    & = \sum\texttt{Softmax}(\frac{\vq\vk^T}{\sqrt{d_k}})\vv.
\end{aligned}
\end{equation}
The attention weights and summations are obtained by the query $\vq$, key $\vk$, and value $\vv$, which are the linear mapping of the context $\vc$ and the object queries $\vphi_q$:
\begin{equation}
    \label{eq:kvq}
    \vq = f_q(\vphi_q);\;\;
    \vk = f_k(\vc);\;\;
    \vv = f_v(\vc).
\end{equation}
Here $f_q, f_k, f_v$ denote the projection for query, key and value in the cross-attention module.

In this work, we aim to pre-train the Transformers of DETR in a self-supervised way, extending the boundary of existing self-supervised pretraining. Motivated by recent advances \cite{chen2020improved, chen2020simple, chen2021exploring, grill2020bootstrap}, we propose Siamese DETR, a Siamese multi-view self-supervised framework designed for Transformers in DETR, in which the model parameters and the learnable object queries in two DETRs are all shared. 
Following UP-DETR \cite{dai2021up}, Siamese DETR aims at learning representations with Transformers in self-supervised pretraining while keeping the backbone model frozen.

\subsection{View Construction}
\label{subsec:views}

\begin{figure}[t!]
    \centering
    \includegraphics[width=0.94\linewidth]{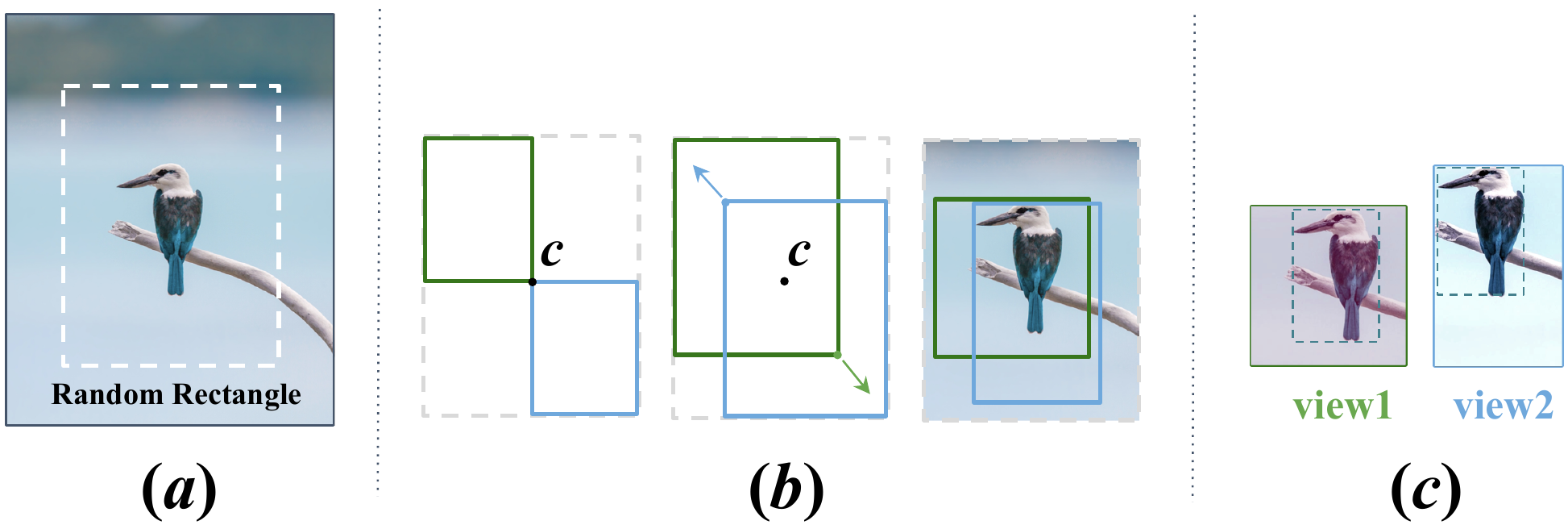}
    \caption{
         View construction using an IoU-constrained policy. \textbf{({\it \textbf{a}})} Generate a random rectangle within the image. \textbf{({\it \textbf{b}})} Generate two sub-rectangles on both sides of the center point $c$. Expand two rectangles along the diagonal while keeping the IoU larger than a threshold $\tau$. Crop two sub-rectangles and apply augmentations. \textbf{({\it \textbf{c}})} Generate offline object proposals in the overlapping area.} 
  \label{fig:view_generation}
\end{figure}

We start with generating two views $\{\vx_1, \vx_2\}$ for each unlabeled image $\vx$, allowing the model to learn view-invariant object-level representations in self-supervised detection pretraining. 
As illustrated in Figure \ref{fig:view_generation}, we introduce an IoU-constrained policy to balance the shared information between two views. 
First, we generate a random rectangle within the image, which covers most content ($50\%$ to $100\%$). 
Then the center point of the rectangle is used as the anchor to create two sub-rectangles. 
By randomly expanding the sub-rectangles along the diagonal, we obtain two rectangles with the IoU larger than a threshold $\tau=0.5$. 
Two rectangles are cropped from the image as the final two views $\{\vx_1, \vx_2\}$.
We further apply randomly and independently sampled transformations on two views $\{\vx_1, \vx_2\}$ following the augmentation pipeline in \cite{chen2021exploring}.
We also apply a box jitter processing following \cite{wei2021aligning} to encourage variance of scales and locations of object proposals across views.

The two augmented views $\{\vx_1, \vx_2\}$ are visually distinct but share adequate semantic content. Following \cite{bar2022detreg, wei2021aligning}, we generate offline object proposals $\vb$ using unsupervised EdgeBoxes \cite{zitnick2014edge} 
in the overlapping area between the two views and randomly select $n=10$ corresponding object proposals in two views $\{\vb_1, \vb_2\}$ from $\vb$. 
These object proposals can provide proper objectness priors to learn object-level representations during pretraining. 
 
\subsection{Multi-view Detection Pretraining}
\label{sec:mvdp}

Given a region extracted from one view, we train the model for answering two questions: 
(1) \textit{where is the corresponding region in another view?} 
(2) \textit{what is the region, \emph{i.e.}, does it semantically similar?} 
In the following, we describe two pretext tasks designed in Siamese DETR: 
(1) learning to locate by \textbf{Multi-View Region Detection} and (2) learning to discriminate by \textbf{Multi-View Semantic Discrimination}. 

\vspace{+1mm}
\noindent\textbf{Preparation.}  
We take the augmented views $\{\vx_1, \vx_2\}$ as inputs to the backbone model to obtain the image-level features: $\{\vh_1, \vh_2\} = \texttt{Backbone}(\{\vx_1, \vx_2\})$.
Then, the object-level region features $\{\vz_1,\vz_2\}$ are extracted from the image-level features $\{\vh_1, \vh_2\}$ based on each object proposal $\{\vb_1,\vb_2\}$ using RoIAlign \cite{he2017mask}:
\begin{equation}
\label{eq:z}
\begin{aligned}
    \vz_1 &= \texttt{RoIAlign}(\vh_1, \vb_1); \\
    \vz_2 &= \texttt{RoIAlign}(\vh_2, \vb_2).
\end{aligned}
\end{equation}
We also obtain the corresponding crop-level region features $\{\vp_1,\vp_2\}\in\mathbb{R}^{C\times H_3\times W_3}$ by first cropping from the augmented views and then extraction:
\begin{equation}
\label{eq:p}
\begin{aligned}
    \vp_1 &= \texttt{Backbone}(\texttt{Crop}(\vx_1, \vb_1)); \\
    \vp_2 &= \texttt{Backbone}(\texttt{Crop}(\vx_2, \vb_2)).
\end{aligned}
\end{equation}
By default, both $\{\vz_1,\vz_2\}$ and $\{\vp_1,\vp_2\}$ are processed with global average pooling before further usage. 
Only the forward pass is involved in preparations, as we focus on pretraining the Transformers.

\vspace{+1mm}
\noindent\textbf{Multi-View Cross-Attention.} 
We propose a Multi-View Cross-Attention ($\texttt{MVCA}$) mechanism that extends the cross-attention module in DETR for multi-view representation learning. 
With the introduced notions for two views $\{(\vc_1, \vz_1, \vp_1, \vb_1), (\vc_2, \vz_2, \vp_2, \vb_2)\} $, we formulate the cross-attention from view $\vx_1$ to view $\vx_2$ as follows:
\begin{equation}
    \label{eq:mvca}
    \widehat{\vq}_2 = \texttt{MVCA}(\vc_2, \vphi_q, \vz_1) = \texttt{MHA}(\vq_1, \vk_2, \vv_2),
\end{equation}
where the query $\vq_1$, key $\vk_2$ and value $\vv_2$ are given by:
\begin{equation}
    \vq_1 = f_q(\vz_1 + \vphi_q);\;\;
    \vk_2 = f_k(\vc_2);\;\;
    \vv_2 = f_v(\vc_2).
    \label{eq:cv_kvq}
\end{equation}
We add the region features $\vz_1$ from view $\vx_1$ to the object queries $\vphi_q$ so that with conditional input of region features $\vz_1$, the object queries $\vphi_q$ can extract the relevant features $\widehat{\vq}_2$ from the global context $\vc_2$ of view $\vx_2$. 
Here, $\widehat{\vq}_2$ is supposed to be aggregated features on view $\vx_2$ that are semantically consistent with the corresponding region features $\vz_1$ on view $\vx_1$.

\vspace{+1mm}
\noindent\textbf{\textit{Learning to locate:}} 
The $\texttt{MVCA}$ mechanism allows us to conduct \textbf{Multi-View Region Detection} directly.
Specifically, with the input of each region feature $\vz_1$ from view $\vx_1$ and its extracted feature $\widehat{\vq}_2$, our goal is to locate the region in view $\vx_2$ that is relative to the region feature $\vz_1$.
We apply a prediction head $f_{box}$ for box prediction:
\begin{equation}
    \label{eq:box_head}
    \widehat{\vb}_2 = f_{box}(\widehat{\vq}_2)\in\mathbb{R}^{N\times 4}.
\end{equation}
After performing bipartite matching \cite{carion2020end}, we calculate the multi-view symmetrical localization loss as:
\begin{equation}
    \label{eq:loc}
    \mathcal{L}_{loc}  = \ell_{box}(\widehat{\vb}_2, \vb_2) + \ell_{box}(\widehat{\vb}_1, \vb_1),
\end{equation}
where $\ell_{box}$ is a combination of generalized IoU loss and $\ell_1$ loss the same as \cite{carion2020end}.

\vspace{+1mm}
\noindent\textbf{\textit{Learning to discriminate:}} Due to the unavailability of semantic label information, we propose \textbf{Multi-View Semantic Discrimination} to learn to discriminate at both global and regional levels. 
First, we apply a prediction head $f_{sem}$ for further discriminative learning:
\begin{equation}
    \label{eq:discriminative_head}
    \widehat{\vp}_2 = f_{sem}(\widehat{\vq}_2)\in\mathbb{R}^{N\times C'}.
\end{equation}
Considering that the context $\vc$ of each view contains global contextual information, we maximize the similarity of the encoded context between two augmented views.
Following \cite{chen2021exploring}, we apply a three-layer MLP (FC-BN-ReLU) and compute the global discrimination loss $\mathcal{L}_{g}$ symmetrically as:
\begin{equation}
\label{eq:disc_g}
\begin{aligned}
    \mathcal{L}_g=\mathcal{C}\big[\texttt{MLP}(\vc_1), &\texttt{detach}(\vc_2)\big] + \\
    &\mathcal{C}\big[\texttt{MLP}(\vc_2), \texttt{detach}(\vc_1)\big],
\end{aligned}
\end{equation}
where $\mathcal{C}$ is the negative cosine similarity.
In addition, due to the semantic consistency between the input region features $\vz_1$ and the extracted features $\widehat{\vq}_2$, we consider maximizing the semantic consistency for each region. 
Here, despite representing the same instances, the crop-level region features $\vp_1$ can provide more discriminative information than object-level region features $\vz_1$. It is because the object-level region features $\vz_1$ are extracted from the image-level features (See Equation \ref{eq:z} and \ref{eq:p}) and contain an aggregate of surrounding contexts with less discriminative information on themselves, especially for small regions. 
It motivates us to replace object-level region features $\vz_1$ with crop-level region features $\vp_1$ as the learning objectives.
Also, to avoid potential training collapse, we reconstruct the crop-level region features $\vp_1$. 
Finally, we formulate the semantic consistency objective to improve the region discrimination as:
\begin{equation}
    \label{eq:disc_r}
    \mathcal{L}_{r} = \mathcal{D}(\widehat{\vp}_2, \vp_1) + \mathcal{D}(\widehat{\vp}_1, \vp_2),
\end{equation}
where $\mathcal{D}$ is the normalized $\ell_2$ distance.

\input{sections/tables/4_1_coco_main_results}

\vspace{+1mm}
\noindent\textbf{Loss Function.} 
Formally, the overall loss function for Siamese DETR is formulated as:
\begin{equation}
    \label{eq:all}
    \mathcal{L} =  \lambda_0\mathcal{L}_{r} + \lambda_1\mathcal{L}_{g} + \lambda_2\mathcal{L}_{loc},
\end{equation}
where $\lambda_{0/1/2}$ are the loss weighting hyper-parameters. 


\subsection{Discussion}

Compared with previous methods \cite{dai2021up, bar2022detreg} and conventional multi-view SSL methods like MoCo \cite{he2020momentum}, we introduce siamese network in the pre-training pipeline, exploring the view-invariant localization ability for DETR pretraining.
Through combining the characteristics of cross-attention with siamese network, Siamese DETR can aggregate the latent information from key view given the cue of the query view, then localizing and discriminating the corresponding regions.  
It is completely different from the common multi-view techniques, \emph{i.e.}, simply contrasting the output of the DETR. 
The proposed Multi-View Cross-Attention enables model to learn to view-invariant localization ability, providing better priors in downstream tasks compared with UP-DETR, which spends most of pre-training time detecting the background classes. 
Meanwhile, DETReg \cite{bar2022detreg}, which can be viewed as detecting the regions generated by proposals and pseudo labels, does not consider view-invariant detection pretraining as well.

%% file: sections/tables/4_1_coco_main_results.tex
\begin{table*}[t!]
    \caption{
        Comparisons of Siamese DETR with supervised/UP-DETR/DETReg in the COCO detection benchmark. The results of all models are achieved by officially-released repositories and pretrained models. Here ``\#epoch'' denotes the number of epochs in downstream finetuning.
    }
    \label{tab:main_result_coco}
    \begin{center}
    \resizebox{0.8\linewidth}{!}{%
        \begin{tabular}{l|c|c|c|c|c|c|c|c|c|c}
        \hline
        Method	              & Backbone & DETR          & $\#$query & $\#$epoch & AP	         & AP$_{50}$     & AP$_{75}$	 & AP$_{s}$      & AP$_{m}$	     & AP$_{l}$      \\ \hline
        \textit{from scratch} & Sup. R50 & Vanilla       & 100       & 150	     & 39.5          & 60.3          & 41.4	         & 17.5	         & 43.0	         & 59.1	         \\
        \textit{from scratch} & SwAV R50 & Vanilla       & 100	     & 150	     & 39.7	         & 60.3	         & 41.7	         & 18.5	         & 43.8	         & 57.5	         \\
        UP-DETR               & SwAV R50 & Vanilla       & 100	     & 150	     & 40.5	         & 60.8          & 42.6	         & 19.0          & 44.4	         & 60.0	         \\
        DETReg                & SwAV R50 & Vanilla       & 100       & 150       & 41.9          & 61.9          & 44.1          & 19.1          & 45.7          & 61.5          \\
        ours                  & SwAV R50 & Vanilla       & 100       & 150	     & \textbf{42.0} & \textbf{63.1} & \textbf{44.2} & \textbf{19.6} & \textbf{46.0} & \textbf{61.9} \\ \hline
        \textit{from scratch} & SwAV R50 & Conditional   & 100	     & 50	     & 37.7	         & 59.6	         & 39.2	         & 17.1	         & 41.7	         & 56.3	         \\
        UP-DETR               & SwAV R50 & Conditional   & 100       & 50	     & 39.4	         & 61.2	         & 41.0	         & 18.1	         & 43.0	         & 58.7	         \\
        DETReg                & SwAV R50 & Conditional   & 100       & 50        & 40.2          & \textbf{61.8} & 42.0          & 19.1          & 43.7          & 60.0          \\
        ours                  & SwAV R50 & Conditional   & 100	     & 50	     & \textbf{40.5} & 61.6          & \textbf{42.6} & \textbf{19.5} & \textbf{44.2} & \textbf{60.1} \\ \hline
        \textit{from scratch} & SwAV R50 & Conditional   & 300	     & 50	     & 41.1	         & 62.3	         & 43.4	         & 20.6	         & 45.0	         & 59.4	         \\
        UP-DETR               & SwAV R50 & Conditional   & 300	     & 50	     & 41.5	         & 63.2	         & 43.6	         & 21.3	         & 45.4	         & 60.2	         \\
        ours                  & SwAV R50 & Conditional   & 300	     & 50	     & \textbf{43.0} & \textbf{64.2} & \textbf{45.6} & \textbf{22.0} & \textbf{47.2} & \textbf{61.8} \\ \hline
        \textit{from scratch} & SwAV R50 & Deform-SS     & 300	     & 50	     & 40.3	         & 60.9	         & 42.9	         & 20.1 	     & 44.8	         & 57.2	         \\
        UP-DETR               & SwAV R50 & Deform-SS     & 300	     & 50	     & 40.8	         & 61.8	         & 43.4	         & 20.4	         & 45.1	         & 59.1          \\
        ours                  & SwAV R50 & Deform-SS     & 300       & 50	     & \textbf{42.1} &\textbf{62.8}  & \textbf{44.7} & \textbf{22.3} & \textbf{46.6} & \textbf{59.9} \\ \hline
        \textit{from scratch} & SwAV R50 & Deform-MS     & 300	     & 50	     & 45.5	         & 64.2	         & 49.4	         & 27.8	         & 49.2	         & 59.4	         \\
        UP-DETR               & SwAV R50 & Deform-MS     & 300	     & 50	     & 45.3	         & 64.5		     & 49.6	         & 26.0	         & 49.2          & 59.9	         \\
        DETReg                & SwAV R50 & Deform-MS     & 300       & 50        & 45.5          & 64.1          & 49.9          & 26.9          & 49.5          & 59.6          \\
        ours                  & SwAV R50 & Deform-MS     & 300       & 50	     & \textbf{46.3} & \textbf{64.6} & \textbf{50.5} & \textbf{28.1} & \textbf{50.1} & \textbf{61.5} \\
        \hline
        \end{tabular}
    }
    \end{center}
\end{table*}

%% file: sections/4_exp.tex
\section{Experiments}

\subsection{Implementation Details}
\label{subsec:impl}

\ifx\arxiv\undefined
    \input{sections/tables/4_2_voc_main_results}
\fi

\noindent\textbf{Architecture.} 
Siamese DETR consists of a frozen ResNet-50 backbone, pretrained by SwAV \cite{swav} on ImageNet, and a Transformer with encoder-decoder architecture. 
Both the Transformer encoder and decoder are stacked with 6 layers of 256 dimensions and 8 attention heads. 
To verify the performance and generalization of our design, we compare Siamese DETR with a baseline model without pretraining (denoted as \textit{from scratch}), UP-DETR \cite{dai2021up}, and DETReg \cite{bar2022detreg}. We use three DETR variants, \emph{i.e.}, original DETR \cite{carion2020end} (denoted as \textit{Vanilla DETR}), Conditional DETR \cite{meng2021conditional}, and Deformable DETR (Single-Scale and Multi-Scale, denoted as Deform-SS and Deform-MS, respectively) \cite{zhu2020deformable}. 
For a fair comparison, the number of object queries is 100 in Vanilla DETR and Conditional DETR.
We also use 300 queries in Conditional DETR and Deformable DETR. 
Besides, we also provide the transfer results of more advanced DETR-like architecture (\emph{e.g.}, DAB-DETR \cite{liu2022dab}) in the appendix.
We implement Siamese DETR based on the MMSelfSup \footnote{https://github.com/open-mmlab/mmselfsup. Apache-2.0 License.}.

\vspace{+1mm}
\noindent\textbf{Evaluation Protocol.} We follow the evaluation protocol in UP-DETR. Specifically, we first pretrain DETR variants on ImageNet ($\sim$1.28 million images) \cite{deng2009imagenet} or COCO \texttt{train2017} ($\sim$118k images) \cite{coco} separately. Then, the ImageNet-pretrained models are finetuned on COCO \texttt{train2017} or PASCAL VOC \texttt{trainval07+12} ($\sim$16.5k images) \cite{voc} separately, while COCO-pretrained models are finetuned on PASCAL VOC \texttt{trainval07+12}. We report COCO-style metrics, including AP, AP$_{50}$, AP$_{75}$, AP$_{s}$, AP$_{m}$, AP$_{l}$, in both COCO \texttt{val2017} and PASCAL VOC \texttt{test2007} benchmarks.

\vspace{+1mm}
\noindent\textbf{Pretraining.}
We use an AdamW~\cite{adamw} optimizer with a total batch size of 256 on ImageNet and 64 on COCO, a learning rate of $1\times10^{-4}$ and a weight decay of $1\times10^{-4}$. 
We adopt a full schedule of 60 epochs, and the learning rate decays at 40 epochs, denoted as the 40/60 schedule for brevity. 
Unless specified, UP-DETR uses random boxes, DETReg uses proposals generated by Selective Search \cite{selective_search}, and Siamese DETR uses Edgeboxes in experiments.

\vspace{+1mm}
\noindent\textbf{Finetuning.}
For Vanilla DETR, we adopt 120/150 and 40/50 schedules in COCO and PASCAL VOC benchmarks.
The initial learning rates of the Transformer and backbone are set to $1\times10^{-4}$ and $5\times10^{-5}$. 
For the other two DETR variants, we report the result under the 40/50 schedule. 
The initial learning rates of the Transformer and backbone are set to $1\times10^{-4}$/$2\times10^{-4}$ and $5\times10^{-5}$/$2\times10^{-5}$ in Conditional/Deformable DETR, respectively.
The batch size of all setups in finetuning is set to 32 for a fair comparison.

\subsection{Main Results}
\label{subsec:main_result}

\noindent\textbf{COCO Object Detection.} 
Table \ref{tab:main_result_coco} shows the transfer results on COCO. Siamese DETR achieves the best performance using three different DETR variants on all setups. 
Especially for Deformable DETR of Multi-Scale, Siamese DETR boosts the model upon baseline more significantly than UP-DETR and DETReg, demonstrating the compatibility of our design with different DETR architectures.

\begin{figure}
  \centering
  \includegraphics[width=0.94\linewidth]{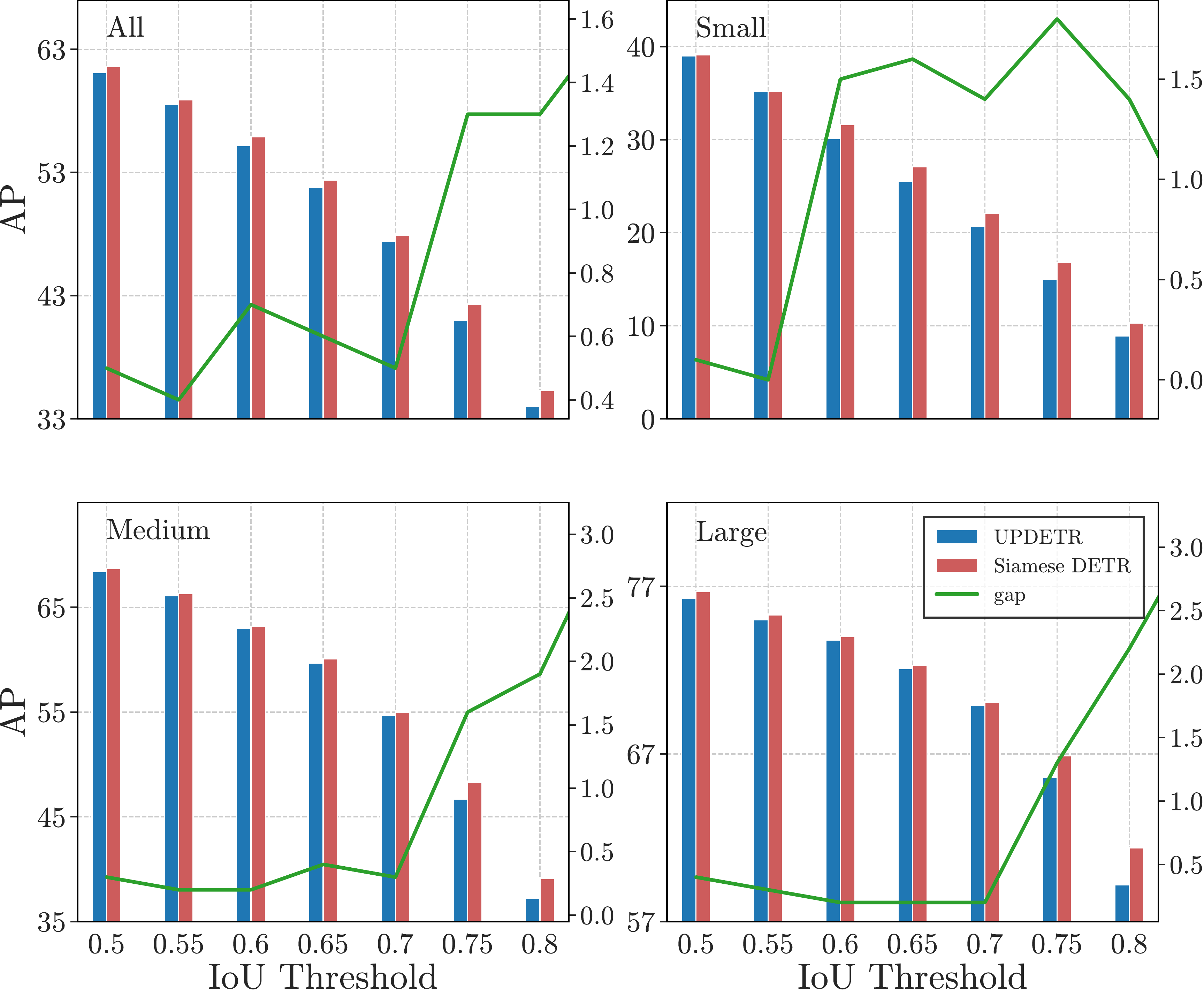}
  \caption{AP on COCO using different IoU Thresholds.}
  \label{fig:ap_iou}
\end{figure}

Besides, when adopting a stricter IoU threshold in metrics, \textit{e.g.}, from AP$_{50}$ to AP$_{75}$, Siamese DETR achieves a more considerable performance lead in most cases. 
It suggests the representations learned by Siamese DETR provide a stronger localization prior.
We further illustrate AP metrics of Siamese DETR and UP-DETR using different IoU thresholds in Figure \ref{fig:ap_iou}. 
Specifically, Siamese DETR performs well in localizing small objects and draws the gap against UP-DETR for medium and large objects when the IoU threshold is greater than 0.7.

\vspace{+1mm}
\noindent\textbf{PASCAL VOC Detection.} 
Table \ref{tab:main_result_voc} shows the transfer results on PASCAL VOC. 
Similar to conclusions on COCO, Siamese DETR achieves the best performance among all approaches on PASCAL VOC. 
We also report the result of COCO-pretrained models. Siamese DETR is 6.4 AP better than UP-DETR and 1.8 AP better than DETReg, which verifies the compatibility of Siamese DETR with different pretraining datasets, especially the scene-centric COCO. 

\subsection{Ablations}
\label{subsec:ablation}

\noindent\textbf{Effectiveness of Two Proposed Pretext Tasks.} 
We train five Siamese DETR variants using Conditional DETR on ImageNet and finetune them on PASCAL VOC. Results are shown in Table \ref{tab:ablation_loss}. We treat UP-DETR as the baseline (56.9 AP on PASCAL VOC), which performs single-view patch detection with Transformer and reconstructs the decoder's output with its input patches. 

\ifx\anonymous\undefined
    \input{sections/tables/4_2_voc_main_results}
\fi

\begin{table}[ht!]
    \caption{Ablations on two proposed pretext tasks, \emph{i.e.}, Multi-View Region Detection and Multi-View Semantic Discrimination. The notation ``R-O'' denotes region discrimination using object-level region features, ``R-C'' denotes region discrimination using crop-level region features, and ``G'' denotes global discrimination.}
    \label{tab:ablation_loss} 
    \begin{center}
        \resizebox{0.80\linewidth}{!}{%
        \begin{tabular}{c|c|c|c}
        \hline
        Method    & Region Det.  & Semantic Disc. & AP            \\ \hline
        UP-DETR   & single-view  & R-O            & 56.9          \\ \hline
        ours (a)  & multi-view   & -              & 57.1          \\
        ours (b)  & multi-view   & G              & 57.3          \\
        ours (c)  & multi-view   & R-O            & 57.3          \\
        ours (d)  & multi-view   & R-C            & 57.7          \\
        ours (e)  & multi-view   & R-C + G        & \textbf{58.1} \\ \hline
        \end{tabular}
        }
    \end{center}
\end{table}

\begin{table}
\hspace{+1mm}
    \caption{
        Comparisons of using different object proposals. The notation ``A$\rightarrow$B'' denotes that the model is pretrained on dataset ``A'' and then finetuned on dataset ``B''.
    }
    \label{tab:main_result_proposal}
    \begin{center}
        \resizebox{0.86\linewidth}{!}{%
        \begin{tabular}{c|c|c|c}
        \hline
        Method      & Dataset                   & Proposals  & AP             \\ \hline
        UP-DETR     & ImageNet$\rightarrow$COCO & Random     & 39.4           \\ 
        DETReg      & ImageNet$\rightarrow$COCO & Random     & 40.3           \\ 
        ours        & ImageNet$\rightarrow$COCO & Random     & \textbf{40.4}  \\ \hline
        UP-DETR     & ImageNet$\rightarrow$COCO & Edgeboxes  & 39.3           \\ 
        DETReg      & ImageNet$\rightarrow$COCO & Edgeboxes  & 40.3           \\ 
        ours        & ImageNet$\rightarrow$COCO & Edgeboxes  & \textbf{40.5}  \\ \hline\hline
        UP-DETR     & COCO$\rightarrow$VOC      & Random     & 51.3           \\ 
        DETReg      & COCO$\rightarrow$VOC      & Random     & 51.9           \\
        ours        & COCO$\rightarrow$VOC      & Random     & \textbf{54.9}  \\ \hline
        DETReg      & COCO$\rightarrow$VOC      & SelectiveSearch  & 55.9           \\ 
        ours        & COCO$\rightarrow$VOC      & SelectiveSearch  & \textbf{56.2}  \\ \hline
        UP-DETR     & COCO$\rightarrow$VOC      & Edgeboxes  & 57.0           \\ 
        DETReg      & COCO$\rightarrow$VOC      & Edgeboxes  & 56.3           \\
        ours        & COCO$\rightarrow$VOC      & Edgeboxes  & \textbf{57.7}  \\ \hline
        \end{tabular}
        }
    \end{center}
\end{table}

By extending single-view detection into a multi-view manner, (a) obtains a competitive result of 57.1 AP, suggesting that the view-invariant representations can perform better in downstream detection tasks. 
We further maximize the multi-view semantic consistency in terms of global and region discrimination, improving the transfer performance by 0.2 AP and 0.6 AP in (b) and (d), respectively. 
Global discrimination (b) brings smaller performance gains than region discrimination (d) in detection tasks, suggesting that it is impractical to directly apply existing instance discrimination pretext tasks \cite{he2020momentum, koch2015siamese} on Transformers of DETR in detection-oriented pretraining tasks. 
We also notice that (d) using crop-level region features achieves better performance than (c) using object-level region features, which verifies more discriminative information in crop-level region features.
Finally, (e) with both Multi-View Region Detection and Multi-View Semantic Discrimination yields the best result of 58.1 AP.

\vspace{+1mm}
\noindent\textbf{Object Proposals.} Edgeboxes \cite{zitnick2014edge} in Siamese DETR provide objectness priors during pretraining. 
We attempt to replace it with boxes generated randomly or by Selective Search.
All experiments are conducted using Conditional DETR. 
The results are shown in Table \ref{tab:main_result_proposal}. 

\begin{figure*}[ht!]
    \centering
    \includegraphics[width=0.78\linewidth]{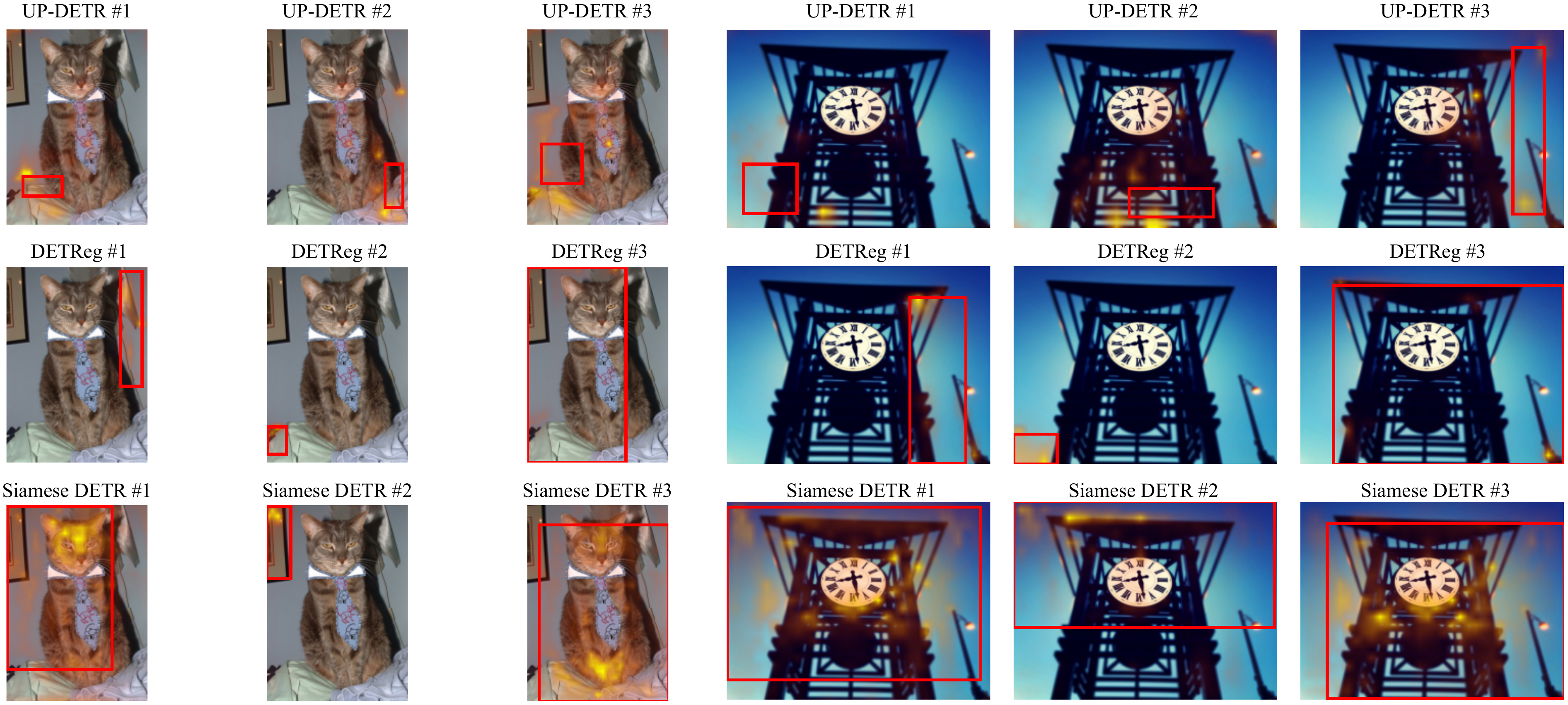}
    \caption{
        Visualization of box predictions and attention maps in downstream tasks.
        All these models (Vanilla DETRs) are initialized by Siamese DETR, UP-DETR, and DETReg without fine-tuning.
        }
  \label{fig:attn_vis_down_3x6}
\end{figure*}

\begin{figure}[t!]
    \centering
    \includegraphics[width=0.95\linewidth]{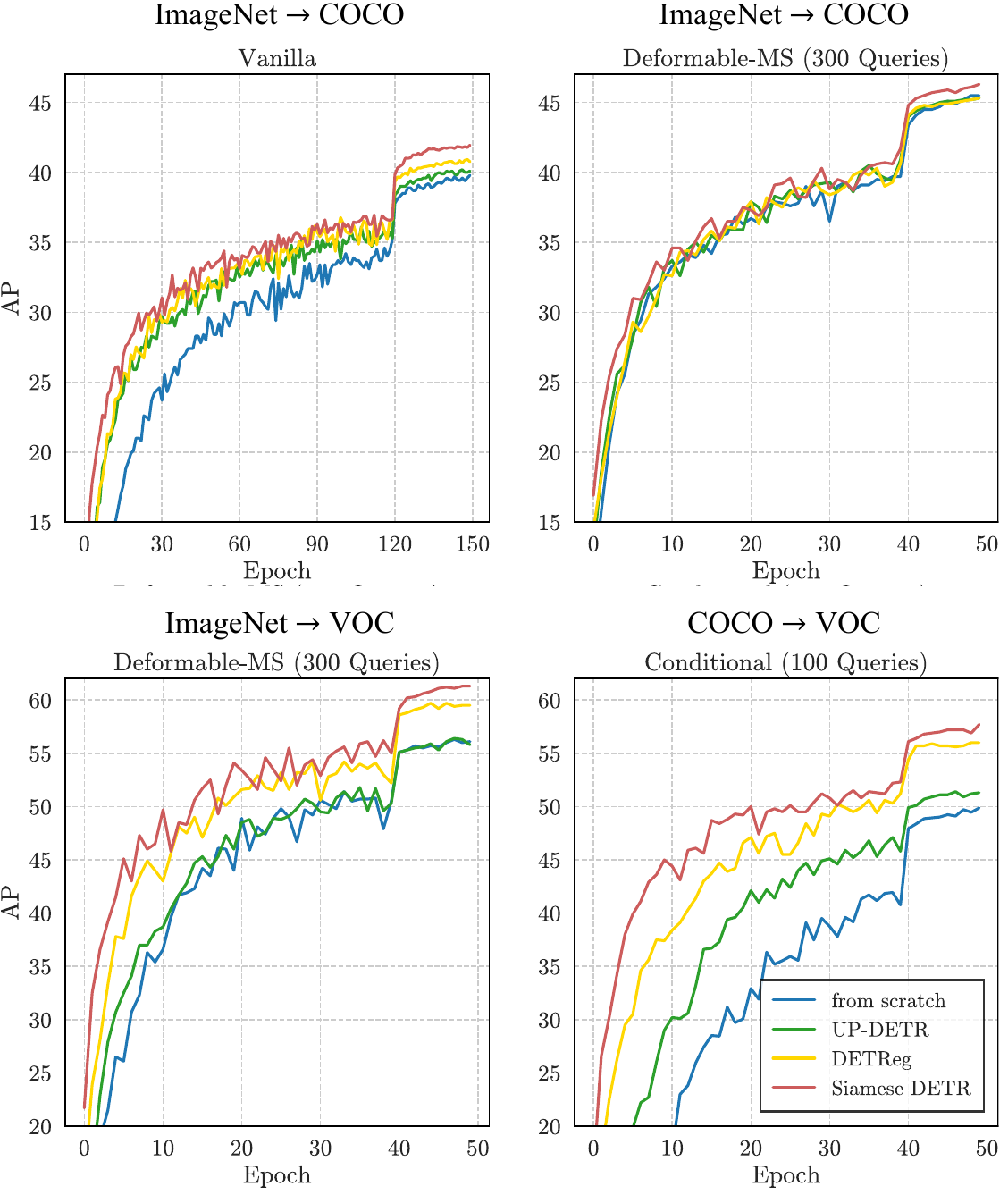}
    \vskip -0.2cm
    \caption{Convergence curves.}
    \label{fig:convergence}
\end{figure}

When pre-trained on object-centric datasets like ImageNet, which contains one single object in the center of the image, better objectness priors bring little improvement. 
In this case, Siamese DETR still outperforms its counterparts by about 0.1 AP using random proposals or Edgeboxes.

When pre-trained on scene-centric datasets like COCO, which contains multiple objects in the image, great improvements are found in all methods after applying better objectness priors from proposals.
Using random proposals, Siamese DETR outperforms UP-DETR by 3.6 AP and DETReg by 3.0 AP in the COCO$\rightarrow$VOC setup, which suggests Siamese DETR learns better detection-oriented representations without any objectness priors. 
When replaced with Selective Search and Edgeboxes, the performance gaps are alleviated. 
In this case, our Siamese DETR with Edgeboxes achieves the best performance among all setups. 

\vspace{+1mm}
\noindent\textbf{Convergence.} The convergence curves of three DETR variants on downstream COCO and PASCAL VOC are illustrated in Figure \ref{fig:convergence}. Compared with UP-DETR, DETReg, and the \textit{from scratch} model, Siamese DETR converges faster and outperforms its counterparts by significant margins. 

\vspace{+1mm}
\noindent\textbf{Visualization.}
We provide qualitative results for further understanding the advantage of Siamese DETR.
In downstream tasks, we use three Vanilla DETRs, initialized by Siamese DETR, UP-DETR, and DETReg pretraining models. 
Figure \ref{fig:attn_vis_down_3x6} illustrates their box predictions and corresponding attention maps of decoder. 
Queries in Siamese DETR have stronger objectness priors, predicting more available box proposals overlapped with objects in the image. 
Benefitting from discriminative representations, cross-attention in Siamese DETR places more focus on the objects in the proposals. 
These qualitative results verify the transferability of Siamese DETR in downstream tasks.

%% file: sections/tables/4_2_voc_main_results.tex
\begin{table*}[t!]
    \caption{
        Comparisons of Siamese DETR with supervised/UP-DETR/DETReg in the PASCAL VOC detection benchmark. The results of all models are achieved by officially-released repositories and pretrained models. Here ``\#epoch'' denotes the number of epochs in downstream finetuning.
    }
    \label{tab:main_result_voc} 
    \begin{center}
        \resizebox{0.90\linewidth}{!}{%
        \begin{tabular}{l | c | c | c | c | c | c | c | c | c | c | c }
        \hline
        Method                & Backbone & DETR	         & Pretrain Dataset & $\#$query & $\#$epoch & AP            & AP$_{50}$     & AP$_{75}$     & AP$_{s}$      & AP$_{m}$      & AP$_{l}$      \\ \hline
        \textit{from scratch} & SwAV R50 & Vanilla       & -                & 100       & 50        & 28.5          & 47.5          & 29.4          & 1.3           & 7.4           & 40.3          \\
        UP-DETR               & SwAV R50 & Vanilla       & ImageNet         & 100       & 50        & 50.0          & 73.5          & 53.4          & 6.5           & 29.5          & 64.1          \\
        DETReg                & SwAV R50 & Vanilla       & ImageNet         & 100       & 50        & 53.8          & 76.5          & 57.3          & \textbf{8.3}  & \textbf{35.4} & 67.5          \\
        ours                  & SwAV R50 & Vanilla       & ImageNet         & 100       & 50        & \textbf{54.4} & \textbf{77.4} & \textbf{57.6} & 7.7           & 35.0          & \textbf{68.4} \\ \hline 
        \textit{from scratch} & SwAV R50 & Vanilla       & -                & 100       & 150       & 47.8          & 73.8          & 50.9          & 5.4           & 27.6          & 61.4          \\
        UP-DETR               & SwAV R50 & Vanilla       & ImageNet         & 100       & 150       & 54.4          & 78.1          & 58.6          & 10.5          & 35.8          & 67.5          \\
        DETReg                & SwAV R50 & Vanilla       & ImageNet         & 100       & 150       & 57.0          & 79.7          & 61.6          & 11.5          & 39.5          & 70.2          \\
        ours                  & SwAV R50 & Vanilla       & ImageNet         & 100       & 150       & \textbf{57.4} & \textbf{80.3} & \textbf{62.2} & \textbf{11.6} & \textbf{39.3} & \textbf{71.0} \\ \hline
        \textit{from scratch} & SwAV R50 & Conditional   & -                & 100  	    & 50	    & 49.9	        & 78.2  	    & 55.3          & 8.1           & 33.5          & 65.1   	    \\
        UP-DETR               & SwAV R50 & Conditional   & ImageNet         & 100 	    & 50	    & 56.9          & 81.5	        & 61.6          & 11.3          & 39.2	        & 69.8	        \\
        DETReg                & SwAV R50 & Conditional   & ImageNet         & 100       & 50        & 57.5          & \textbf{82.0} & 62.4          & 11.8          & \textbf{40.7} & 71.0          \\
        ours                  & SwAV R50 & Conditional   & ImageNet         & 100 	    & 50	    & \textbf{58.1} & 81.6          & \textbf{62.8} & \textbf{12.2} & 40.6          & \textbf{71.5} \\ \hline 
        \textit{from scratch} & SwAV R50 & Deform-SS     & -                & 300 	    & 50        & 53.8	        & 79.5	        & 59.1	        & 11.8	        & 39.8	        & 65.7          \\
        UP-DETR               & SwAV R50 & Deform-SS     & ImageNet         & 300       & 50        & 54.0	        & 79.3	        & 58.8          & 11.4	        & 38.6	        & 66.1          \\
        ours                  & SwAV R50 & Deform-SS     & ImageNet         & 300       & 50        &\textbf{58.0}  & \textbf{81.8} & \textbf{64.0} & \textbf{14.0} & \textbf{43.3} & \textbf{70.3} \\ \hline 
        \textit{from scratch} & SwAV R50 & Deform-MS     & -                & 300 	    & 50        & 56.1          & 80.7          & 61.9          & 17.4          & 42.7          & 66.4          \\
        UP-DETR               & SwAV R50 & Deform-MS     & ImageNet         & 300       & 50        & 56.4          & 80.9          & 62.3          & 17.3          & 41.3          & 67.4          \\
        DETReg                & SwAV R50 & Deform-MS     & ImageNet         & 300       & 50        &59.7           & 82.0          & 66.4          & 18.2          & 46.4          & 70.4          \\
        ours                  & SwAV R50 & Deform-MS     & ImageNet         & 300       & 50        & \textbf{61.2} & \textbf{82.9} & \textbf{67.7} & \textbf{19.3} & \textbf{47.1} & \textbf{72.2} \\ \hline 
        \textit{from scratch} & SwAV R50 & Conditional   & -                & 100 	    & 50	    & 49.9          & 78.2          & 55.3          & 8.1           & 33.5          & 65.1          \\
        UP-DETR               & SwAV R50 & Conditional   & COCO             & 100       & 50        & 51.3          & 79.0          & 55.3          & 9.5           & 35.3          & 63.7          \\
        DETReg                & SwAV R50 & Conditional   & COCO             & 100       & 50        & 55.9          & 80.0          & 61.6          & \textbf{11.0} & 39.3          & 68.5          \\
        ours                  & SwAV R50 & Conditional   & COCO             & 100       & 50	    & \textbf{57.7} & \textbf{80.9} & \textbf{62.5} & \textbf{11.0} & \textbf{40.4} & \textbf{70.9} \\
        \hline
        \end{tabular}
        }
    \end{center}
\end{table*}

%% file: sections/5_conclusion.tex
\section{Conclusion and Limitations}

In this paper, we propose Siamese DETR, a novel self-supervised pretraining method for DETR. 
With two newly-designed pretext tasks, we directly locate the query regions in a cross-view manner and maximize multi-view semantic consistency, learning localization and discrimination representations transfer to downstream detection tasks.
Siamese DETR achieves better performance with three DETR variants in COCO and PASCAL VOC benchmark against its counterpart. 
Despite the great potential for pretraining DETR, Siamese DETR has a limitation in that it still relies on a pre-trained CNN, \textit{e.g.}, SwAV, without integrating CNN and Transformer into a unified pretraining paradigm. 
In our future work, a more efficient framework for the end-to-end DETR pretraining is desirable.

%% file: sections/6_acknownledgement.tex
\vspace{+1mm}
\noindent\textbf{Acknowledgement.} 
This study is supported by National Key Research and Development Program of China (2021YFB1714300). It is also supported under the RIE2020 Industry Alignment Fund - Industry Collaboration Projects (IAF-ICP) Funding Initiative, National Natural Science Foundation of China (62132001), Singapore MOE AcRF Tier 2 (MOE-T2EP20120-0001), as well as cash and in-kind contribution from the industry partner(s).

%% file: sections/7_appendix.tex
\appendix

\section{More Details of DETR}
\label{sec:more_details_detr}

\subsection{Multi-head Attention}
\label{subsec:mha}

\noindent\textbf{Single-head Attention} (\texttt{SHA}) \textbf{.} 
We start with the attention mechanism with single head. 
Given the key-value sequence $\vx_{kv}$, query sequence $\vx_q$, and linear projection of the attention head $f_v, f_k, f_q$, we can compute so-called query $\vq$, key $\vk$, and value $\vv$ embeddings:
\begin{equation}
\begin{aligned}
    \vq&=f_q(\vx_q+\vphi_q);\\
    \vk&=f_k(\vx_{kv}+\vphi_p);\\
    \vv&=f_v(\vx_{kv}),
\end{aligned}
\end{equation}
where $\vphi_{p}$ is the positional embedding for the key-value sequences, and $\vphi_{q}$ is the positional embedding for the query sequences.
And the attention outputs $\widehat{q}$ are computed by the aggregation of weighted values:
\begin{equation}
    \widehat{q}=\texttt{SHA}(\vq,\vk,\vv)=\sum_j\alpha_{i,j}\vv_j,
\end{equation}
where the attention weights is based on softmax of scaled dot products between $i$-th query and $j$-th key:
\begin{equation}
    \alpha_{i,j}=\texttt{Softmax}(\frac{\vq_i\vk_j^T}{\sqrt{d_k}}),
\end{equation}
where $d_k$ is a scaling factor.

\vspace{+1mm}
\noindent\textbf{Multi-head Attention} (\texttt{MHA}) \textbf{.}
Through concatenating $N$ single-head attentions followed by a projection $f_{\texttt{MHA}}$, we can compute the multi-head attention:
\begin{equation}
    \begin{aligned}
        \widehat{q}&=\texttt{MHA}(\vq,\vk,\vv)\\
        &=f_{\texttt{MHA}}\Big(\texttt{Concat}\big[\texttt{SHA}_0(\vq,\vk,\vv),\dots,\texttt{SHA}_N(\vq,\vk,\vv)\big]\Big).
    \end{aligned}
\end{equation}
Note that the output $\widehat{q}$ is the same size as the input query sequences $\vx_q$. 

\subsection{Bipartite Matching}
\label{subsec:bipartite}

Following \cite{carion2020end, dai2021up}, we apply the Hungarian algorithm \cite{hungarian} to match the $N$ predictions $\widehat{y}$ with the ground truth $y$. The matching loss $\mathcal{H}$ is defined as:
\begin{equation}
  \mathcal{H}(y, \widehat{y})=\sum_{i=1}^N\Big[-\eta_0\log\widehat{\vk}_{\widehat{\sigma}(i)}+\1_\mathrm{\{\vk_i=1\}}\mathcal{L}_{box}(\vb_i,\widehat{\vb}_{\widehat{\sigma}(i)})\Big],
  \label{eq:hungarian}
\end{equation}
where $\vk$ is the binary classification indicating whether each query is matched ($\vk_i=1$) or not ($\vk_i=0$), $\mathcal{L}_{box}$ is a combination of generalized IoU loss \cite{rezatofighi2019generalized} and $\ell_1$ loss, and $\widehat{\sigma}(i)$ is the index of prediction that matching with $i$-th ground truth optimally. The coefficients of binary classification $\eta_0$, generalized IoU loss $\eta_1$, and $\ell_1$ loss $\eta_2$ in Equation \ref{eq:hungarian} are set to 1, 2, 5 following \cite{carion2020end}, respectively.

\section{More Ablations}

\begin{figure}
    \centering
    \includegraphics[width=\linewidth]{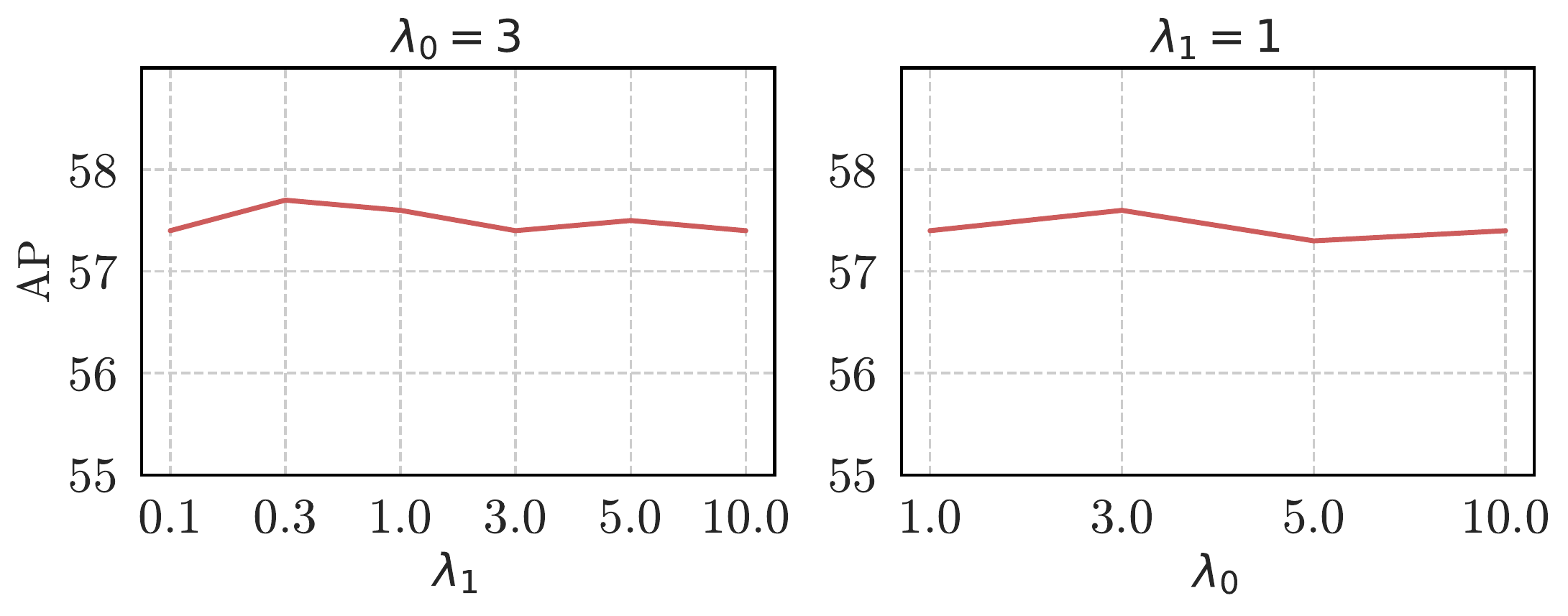}
    \caption{Sensitivity of Hyper-parameters in Siamese DETR.}
    \label{fig:hyperparameter_sensitivity}
\end{figure}

\noindent\textbf{Hyper-parameters.}
We follow \cite{carion2020end} to set loss weight of $\mathcal{L}_{loc}$ ($\lambda_2$) to 1.0 in all setups and further ablate the $\lambda_0$ and $\lambda_1$ using Conditional DETR on COCO.
Figure \ref{fig:hyperparameter_sensitivity} illustrates the sensitivity of  $\lambda_0$ and $\lambda_1$. It suggests that the transfer performance is robust to $\lambda_0$ and $\lambda_1$ variation. To yield the best performance, we set $\lambda_0$, $\lambda_1$ to 3, 10 on ImageNet, and $\lambda_0$, $\lambda_1$ to 0.3, 3 on COCO. 
For other novel datasets, a simple selection (\textit{e.g.}, $\lambda_1=1.0,\lambda_2=1.0$) will be okay.

\begin{figure}[h]
    \centering
    \includegraphics[width=0.7\linewidth]{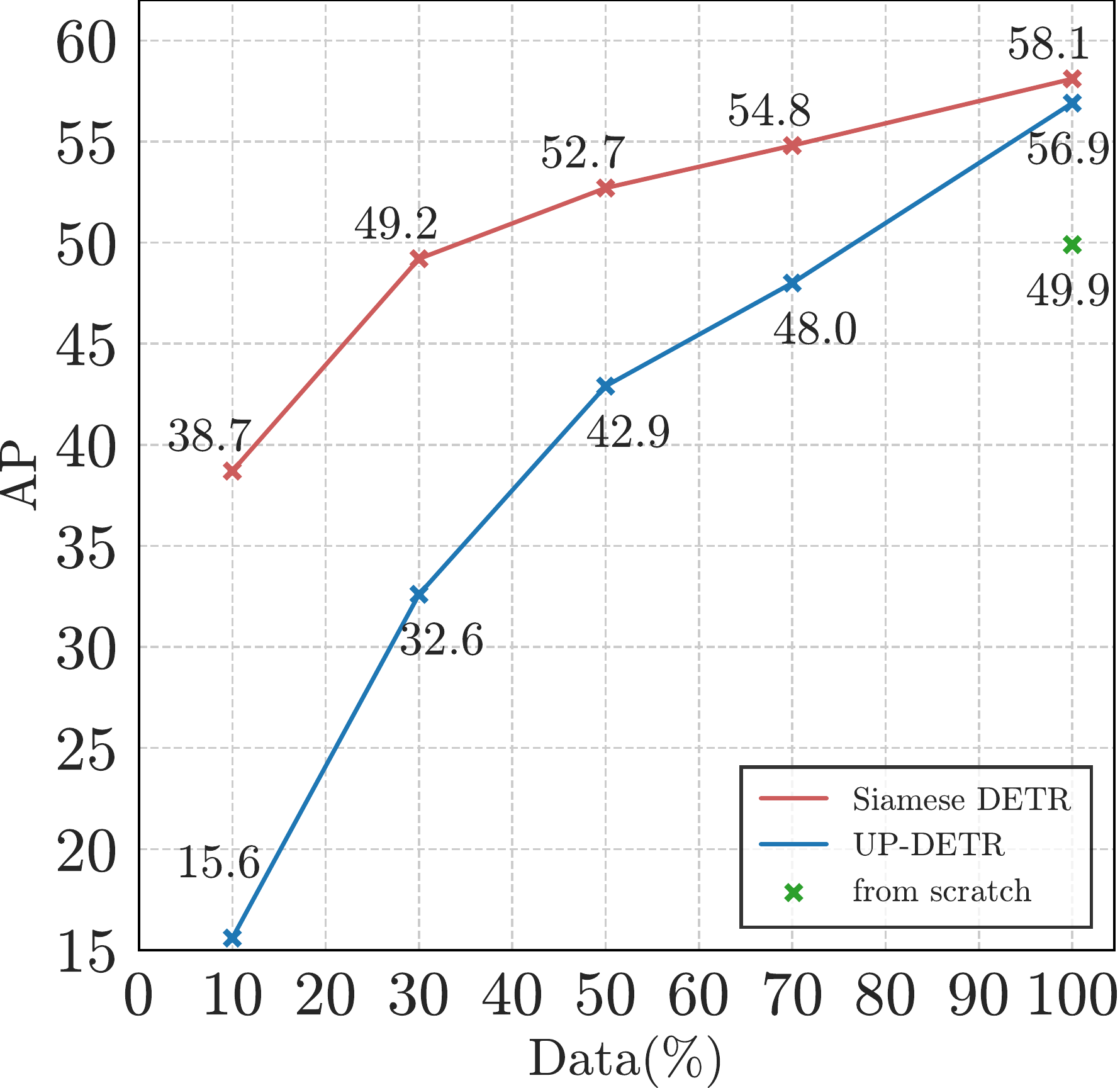}
    \caption{
        Data efficiency of Siamese DETR. We finetune Siamese DETR and UP-DETR using 10\%/30\%/50\%/70\% PASCAL VOC datasets.}
    \label{fig:dataeff}
\end{figure}

\vspace{+1mm}
\noindent\textbf{Data Efficiency.}
Self-supervised pretrained models not only achieve high performance when transferring to downstream tasks, but provide a better initialization when using limited data. To verify the data efficiency of Siamese DETR, we consider the transfer performance on the limited amount of downstream datasets.
Specifically, we pretrain Conditional DETR on ImageNet and finetune it on 10\%/30\%/50\%/70\% PASCAL VOC datasets. 
All these splits are selected randomly. 
As shown in Figure \ref{fig:dataeff}, Siamese DETR can achieve a similar (49.3 AP) performance with from scratch model (49.9) using only 30\% of datasets. Moreover, Siamese DETR outperforms UP-DETR by a large margin in all splits.

\begin{table}
    \caption{Ablations on downstream initialization. We initialize the Vanilla DETR using models pretrained by Siamese DETR, DETReg and UP-DETR without finetining. We report average recall with detecting top $K$ objects, denoted as AR@$K$.}
    \label{tab:init}
    \begin{center}
        \resizebox{0.65\linewidth}{!}{
        \begin{tabular}{c|c|c|c}
        \toprule
        Method       & AR@1 & AR@10 & AR@100 \\ \hline
        random       & 0.0  & 0.1   & 0.5    \\
        UP-DETR      & 9.5  & 17.9  & 24.0   \\
        DETReg       & 11.2 & 20.5  & 26.5   \\ 
        ours         & 12.4 & 23.0  & 30.7   \\ 
        \bottomrule
        \end{tabular}
        }
    \end{center}
\end{table}

\vspace{+1mm}
\noindent\textbf{Downstream initialization.}
To investigate the downstream initialization of Siamese DETR, we pretrain the Vanilla DETR on ImageNet and only finetune the box prediction and classifier head on PASCAL VOC while keeping the parameters of pretrained CNN backbone, encoder and decoder fixed. 
We report average recall with detecting top $K$ objects, denoted as AR@$K$.
As shown in Table \ref{tab:init}, Siamese DETR outperforms its counterparts and random initialization. 

\vspace{+1mm}
\noindent\textbf{More DETR-like architecture.} 
We provide transfer results of Siamese DETR with more advanced DETR-like architecture, \emph{i.e.}, DAB-Deformable-DETR with 300 queries \cite{liu2022dab} and DN-DAB-Deformable-DETR with 300 queries \cite{li2022dn}.
We follow the default setup in their origin paper. 
The results are shown in Table \ref{tab:more_detr_results}.
Both DAB-DETR and DN-DETR can benefit from the initialization of Siamese DETR, verifying the generalization of Siamese DETR. 
Furthermore, Siamese DETR also leads a significant margin when using limited downstream datasets.

\begin{table}
    \caption{More DETR-like architecture. We pretrain the model using Siamese DETR on COCO for the 40/60 schedule, then finetune on full/10\% PASCAL VOC dataset.}
    \label{tab:more_detr_results}
    \begin{center}
        \resizebox{\linewidth}{!}{
        \begin{tabular}{c|c|c|c|c}
        \toprule
        \multirow{2}{*}{DETR} & \multicolumn{2}{c|}{VOC}                    & \multicolumn{2}{c}{VOC 10\%}                  \\ 
                              & DAB-DETR             & DN-DETR              & DAB-DETR              & DN-DETR               \\ \hline
        \textit{from scratch} & 57.9                 & 58.9                 & 32.2                  & 32.9                  \\ \hline
        Siamese DETR          & \textbf{62.2 (+4.3)} & \textbf{63.4 (+4.5)} & \textbf{41.8 (+9.6)}  & \textbf{43.6 (+10.7)} \\ 
        \bottomrule   
        \end{tabular}
        }
    \end{center}
\end{table}

\vspace{+1mm}
\noindent\textbf{Computation Cost.}
Firstly, it is emphasised that none of three pre-training methods increase extra cost in GPU memory and time during downstream finetuning. We provide a quantitative comparison on GPU memory cost, number of trainable parameters and iteration time during pretraining in Table \ref{tab:mem_iter}. All three methods follow the same setup, \emph{i.e.}, pretraining Conditional DETR (100 queries) on COCO using the same GPU. The batch size is set to 4 on each GPU. The input images  are processed with the same augmentation.

Compared with UP-DETR on GPU memory cost and number of parameters, there is a slight increase in Siamese DETR because the parameters in Siamese DETR are all shared. Besides, performing multi-view learning and adding crop-level features does not bring too much time cost.

\begin{table}
    \caption{Memory cost and iteration time during pretraining. }
    \label{tab:mem_iter}
    \begin{center}
        \resizebox{0.9\linewidth}{!}{
        \begin{tabular}{c|c|c|c}
        \toprule
        Method       & GPU Mem. & \#Params & Iteration time \\ \hline
        Siamese DETR & 7528 MB  & 19.49    & 0.4036 s/it    \\ 
        UP-DETR      & 7377 MB  & 19.23    & 0.3118 s/it    \\ 
        DETReg       & 8885 MB  & 19.23    & 0.3528 s/it    \\ 
        \bottomrule
        \end{tabular}
        }
    \end{center}
\end{table}

\section{More Visualization}

\subsection{Convergence}
\label{sec:more_curves}

\begin{figure*}
    \centering
    \includegraphics[width=0.9\linewidth]{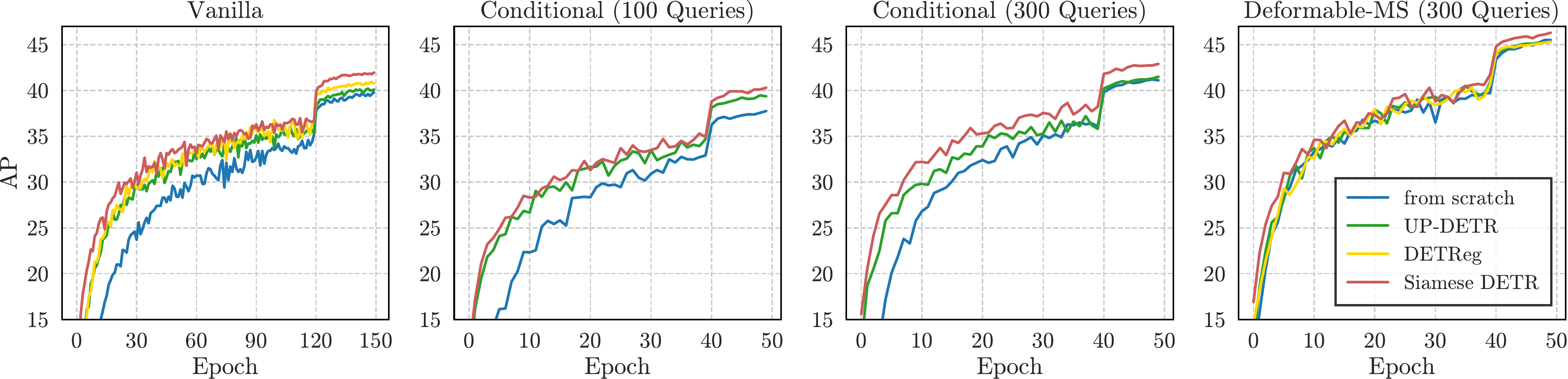}
    \caption{
        Illustration of convergence curves when finetuned on COCO.}
    \label{fig:coco_curves}
\end{figure*}

\begin{figure*}
    \centering
    \includegraphics[width=0.9\linewidth]{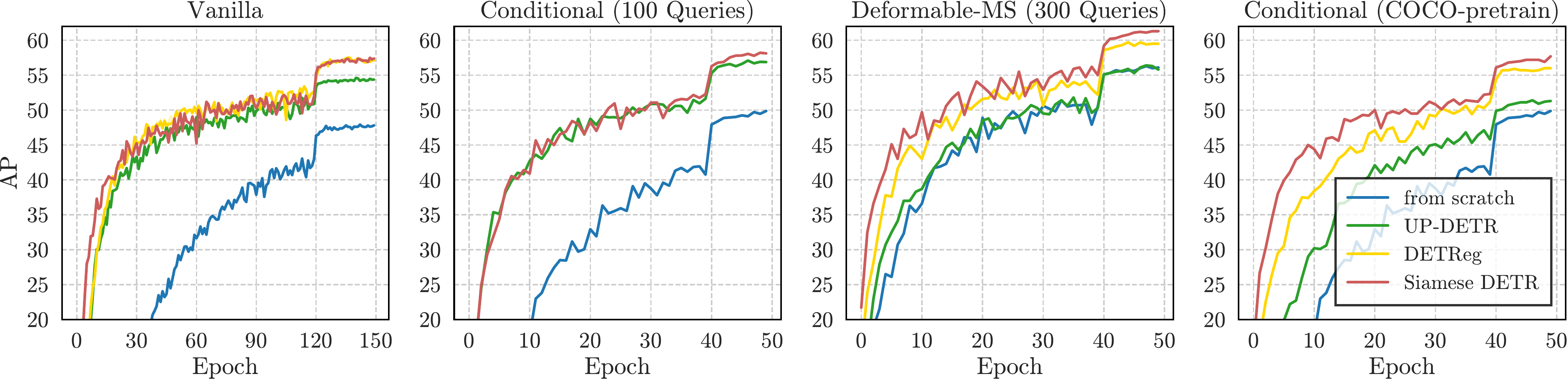}
    \caption{
        Illustration of convergence curves when finetuned on PASCAL VOC.}
    \label{fig:voc_curves}
\end{figure*}

Figure \ref{fig:coco_curves} and Figure \ref{fig:voc_curves} illustrate the convergence curves of models finetuned on COCO and PASCAL VOC, respectively. The model initialized by Siamese DETR converges faster and outperforms its counterparts by significant margins in all setups.

\subsection{More Qualitative Results}
\label{sec:more_vis}

\begin{figure*}
    \centering
    \includegraphics[width=0.9\linewidth]{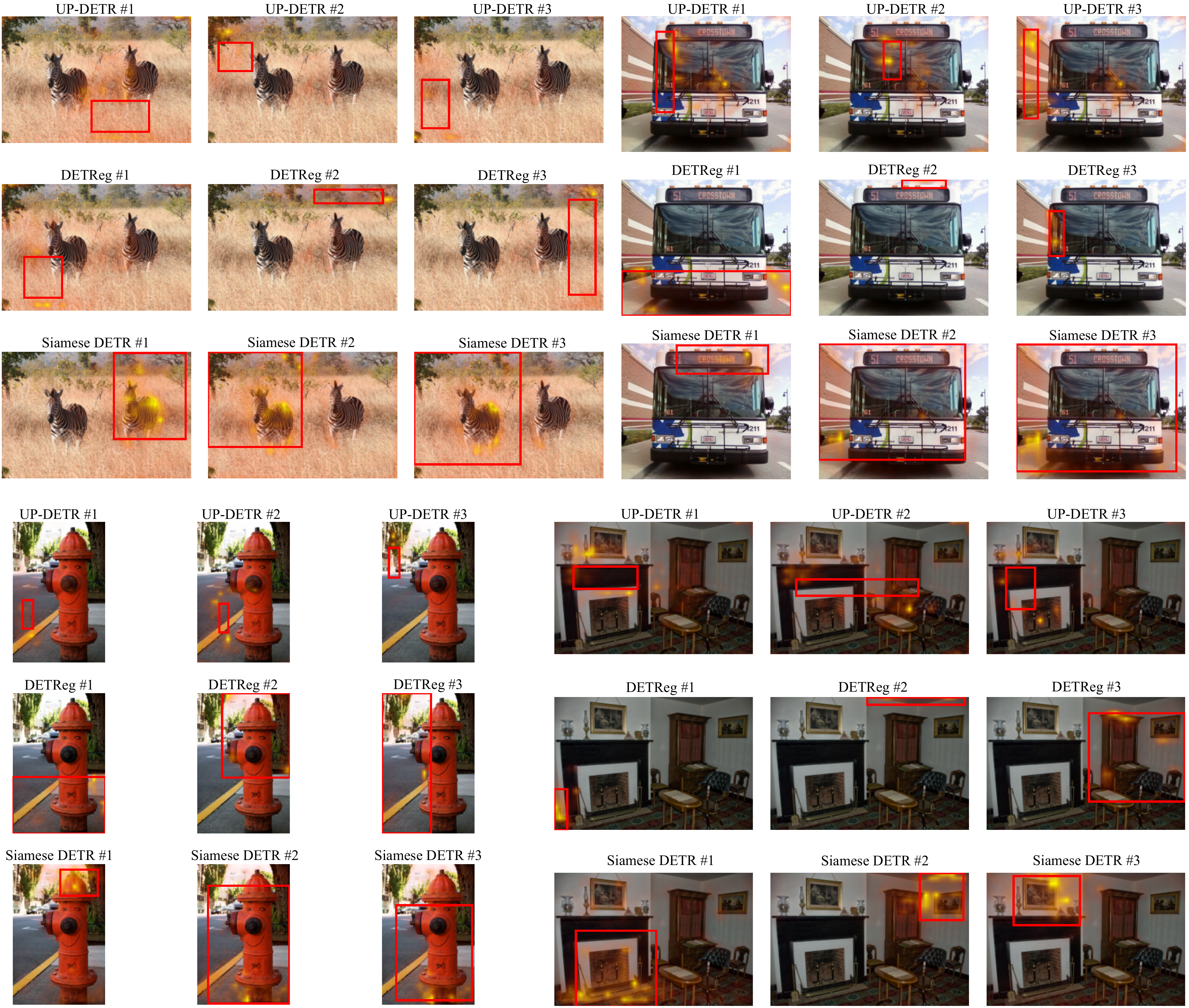}
    \caption{
        More visualization on box predictions and attention maps when initializing the downstream models using Siamese DETR, UP-DETR and DETReg without finetuning.}
    \label{fig:more_vis}
\end{figure*}

We also provide more qualitative results of box predictions and corresponding attention maps when initializing the downstream model using Siamese DETR, UP-DETR, and DETReg without finetuning in Figure \ref{fig:more_vis}. The visualization results verify better transferability of Siamese DETR against its counterpart.

%% file: arxiv.bbl
\begin{thebibliography}{10}\itemsep=-1pt

\bibitem{bar2022detreg}
Amir Bar, Xin Wang, Vadim Kantorov, Colorado~J Reed, Roei Herzig, Gal Chechik,
  Anna Rohrbach, Trevor Darrell, and Amir Globerson.
\newblock Detreg: Unsupervised pretraining with region priors for object
  detection.
\newblock In {\em Proceedings of the IEEE/CVF Conference on Computer Vision and
  Pattern Recognition}, pages 14605--14615, 2022.

\bibitem{bromley1993signature}
Jane Bromley, Isabelle Guyon, Yann LeCun, Eduard S{\"a}ckinger, and Roopak
  Shah.
\newblock Signature verification using a ``{S}iamese'' time delay neural
  network.
\newblock {\em NeurIPS}, 6:737--744, 1993.

\bibitem{carion2020end}
Nicolas Carion, Francisco Massa, Gabriel Synnaeve, Nicolas Usunier, Alexander
  Kirillov, and Sergey Zagoruyko.
\newblock End-to-end object detection with transformers.
\newblock In {\em eccv}, pages 213--229. Springer, 2020.

\bibitem{caron2020unsupervised}
Mathilde Caron, Ishan Misra, Julien Mairal, Priya Goyal, Piotr Bojanowski, and
  Armand Joulin.
\newblock Unsupervised learning of visual features by contrasting cluster
  assignments.
\newblock {\em NeurIPS}, 2020.

\bibitem{swav}
Mathilde Caron, Ishan Misra, Julien Mairal, Priya Goyal, Piotr Bojanowski, and
  Armand Joulin.
\newblock Unsupervised learning of visual features by contrasting cluster
  assignments.
\newblock 2020.

\bibitem{chen2020simple}
Ting Chen, Simon Kornblith, Mohammad Norouzi, and Geoffrey Hinton.
\newblock A simple framework for contrastive learning of visual
  representations.
\newblock In {\em ICML}, pages 1597--1607. PMLR, 2020.

\bibitem{chen2020improved}
Xinlei Chen, Haoqi Fan, Ross Girshick, and Kaiming He.
\newblock Improved baselines with momentum contrastive learning.
\newblock {\em arXiv}, 2020.

\bibitem{chen2021exploring}
Xinlei Chen and Kaiming He.
\newblock Exploring simple siamese representation learning.
\newblock In {\em CVPR}, pages 15750--15758, 2021.

\bibitem{dai2021up}
Zhigang Dai, Bolun Cai, Yugeng Lin, and Junying Chen.
\newblock Up-detr: Unsupervised pre-training for object detection with
  transformers.
\newblock In {\em CVPR}, pages 1601--1610, 2021.

\bibitem{deng2009imagenet}
Jia Deng, Wei Dong, Richard Socher, Li-Jia Li, Kai Li, and Li Fei-Fei.
\newblock Imagenet: A large-scale hierarchical image database.
\newblock In {\em CVPR}, pages 248--255. Ieee, 2009.

\bibitem{dosovitskiy2020image}
Alexey Dosovitskiy, Lucas Beyer, Alexander Kolesnikov, Dirk Weissenborn,
  Xiaohua Zhai, Thomas Unterthiner, Mostafa Dehghani, Matthias Minderer, Georg
  Heigold, Sylvain Gelly, et~al.
\newblock An image is worth 16x16 words: Transformers for image recognition at
  scale.
\newblock {\em ICLR}, 2020.

\bibitem{voc}
Mark Everingham, Luc Van~Gool, Christopher~KI Williams, John Winn, and Andrew
  Zisserman.
\newblock The pascal visual object classes (voc) challenge.
\newblock {\em IJCV}, 88(2):303--338, 2010.

\bibitem{Gao_2021_ICCV}
Peng Gao, Minghang Zheng, Xiaogang Wang, Jifeng Dai, and Hongsheng Li.
\newblock Fast convergence of detr with spatially modulated co-attention.
\newblock In {\em ICCV}, pages 3621--3630, October 2021.

\bibitem{grill2020bootstrap}
Jean-Bastien Grill, Florian Strub, Florent Altch{\'e}, Corentin Tallec,
  Pierre~H Richemond, Elena Buchatskaya, Carl Doersch, Bernardo~Avila Pires,
  Zhaohan~Daniel Guo, Mohammad~Gheshlaghi Azar, et~al.
\newblock Bootstrap your own latent: A new approach to self-supervised
  learning.
\newblock {\em NeurIPS}, 2020.

\bibitem{he2020momentum}
Kaiming He, Haoqi Fan, Yuxin Wu, Saining Xie, and Ross Girshick.
\newblock Momentum contrast for unsupervised visual representation learning.
\newblock In {\em CVPR}, pages 9729--9738, 2020.

\bibitem{he2017mask}
Kaiming He, Georgia Gkioxari, Piotr Doll{\'a}r, and Ross Girshick.
\newblock Mask r-cnn.
\newblock In {\em ICCV}, pages 2961--2969, 2017.

\bibitem{he2016deep}
Kaiming He, Xiangyu Zhang, Shaoqing Ren, and Jian Sun.
\newblock Deep residual learning for image recognition.
\newblock In {\em CVPR}, pages 770--778, 2016.

\bibitem{Henaff_2021_ICCV}
Olivier~J. H\'enaff, Skanda Koppula, Jean-Baptiste Alayrac, Aaron van~den Oord,
  Oriol Vinyals, and Jo\~ao Carreira.
\newblock Efficient visual pretraining with contrastive detection.
\newblock In {\em ICCV}, pages 10086--10096, October 2021.

\bibitem{koch2015siamese}
Gregory Koch, Richard Zemel, Ruslan Salakhutdinov, et~al.
\newblock Siamese neural networks for one-shot image recognition.
\newblock In {\em ICMLW}, volume~2. Lille, 2015.

\bibitem{hungarian}
Harold~W Kuhn.
\newblock The hungarian method for the assignment problem.
\newblock {\em Naval research logistics quarterly}, 2(1-2):83--97, 1955.

\bibitem{li2022dn}
Feng Li, Hao Zhang, Shilong Liu, Jian Guo, Lionel~M Ni, and Lei Zhang.
\newblock Dn-detr: Accelerate detr training by introducing query denoising.
\newblock In {\em Proceedings of the IEEE/CVF Conference on Computer Vision and
  Pattern Recognition}, pages 13619--13627, 2022.

\bibitem{li2021prototypical}
Junnan Li, Pan Zhou, Caiming Xiong, and Steven~CH Hoi.
\newblock Prototypical contrastive learning of unsupervised representations.
\newblock {\em ICLR}, 2021.

\bibitem{coco}
Tsung-Yi Lin, Michael Maire, Serge Belongie, James Hays, Pietro Perona, Deva
  Ramanan, Piotr Doll{\'a}r, and C~Lawrence Zitnick.
\newblock Microsoft coco: Common objects in context.
\newblock In {\em ECCV}, pages 740--755. Springer, 2014.

\bibitem{liu2022dab}
Shilong Liu, Feng Li, Hao Zhang, Xiao Yang, Xianbiao Qi, Hang Su, Jun Zhu, and
  Lei Zhang.
\newblock Dab-detr: Dynamic anchor boxes are better queries for detr.
\newblock {\em arXiv preprint arXiv:2201.12329}, 2022.

\bibitem{liu2021unbiased}
Yen-Cheng Liu, Chih-Yao Ma, Zijian He, Chia-Wen Kuo, Kan Chen, Peizhao Zhang,
  Bichen Wu, Zsolt Kira, and Peter Vajda.
\newblock Unbiased teacher for semi-supervised object detection.
\newblock {\em arXiv preprint arXiv:2102.09480}, 2021.

\bibitem{adamw}
Ilya Loshchilov and Frank Hutter.
\newblock Decoupled weight decay regularization.
\newblock {\em arXiv}, 2017.

\bibitem{meng2021conditional}
Depu Meng, Xiaokang Chen, Zejia Fan, Gang Zeng, Houqiang Li, Yuhui Yuan, Lei
  Sun, and Jingdong Wang.
\newblock Conditional detr for fast training convergence.
\newblock In {\em ICCV}, pages 3651--3660, 2021.

\bibitem{rezatofighi2019generalized}
Hamid Rezatofighi, Nathan Tsoi, JunYoung Gwak, Amir Sadeghian, Ian Reid, and
  Silvio Savarese.
\newblock Generalized intersection over union: A metric and a loss for bounding
  box regression.
\newblock In {\em CVPR}, pages 658--666, 2019.

\bibitem{taigman2014deepface}
Yaniv Taigman, Ming Yang, Marc'Aurelio Ranzato, and Lior Wolf.
\newblock Deepface: Closing the gap to human-level performance in face
  verification.
\newblock In {\em CVPR}, pages 1701--1708, 2014.

\bibitem{selective_search}
Jasper~RR Uijlings, Koen~EA Van De~Sande, Theo Gevers, and Arnold~WM Smeulders.
\newblock Selective search for object recognition.
\newblock {\em IJCV}, 104(2):154--171, 2013.

\bibitem{vaswani2017attention}
Ashish Vaswani, Noam Shazeer, Niki Parmar, Jakob Uszkoreit, Llion Jones,
  Aidan~N Gomez, {\L}ukasz Kaiser, and Illia Polosukhin.
\newblock Attention is all you need.
\newblock In {\em NeurIPS}, pages 5998--6008, 2017.

\bibitem{wang2021dense}
Xinlong Wang, Rufeng Zhang, Chunhua Shen, Tao Kong, and Lei Li.
\newblock Dense contrastive learning for self-supervised visual pre-training.
\newblock In {\em CVPR}, pages 3024--3033, 2021.

\bibitem{wei2021aligning}
Fangyun Wei, Yue Gao, Zhirong Wu, Han Hu, and Stephen Lin.
\newblock Aligning pretraining for detection via object-level contrastive
  learning.
\newblock {\em arXiv}, 2021.

\bibitem{xie2021detco}
Enze Xie, Jian Ding, Wenhai Wang, Xiaohang Zhan, Hang Xu, Peize Sun, Zhenguo
  Li, and Ping Luo.
\newblock Detco: Unsupervised contrastive learning for object detection.
\newblock In {\em ICCV}, pages 8392--8401, 2021.

\bibitem{xie2021unsupervised}
Jiahao Xie, Xiaohang Zhan, Ziwei Liu, Yew~Soon Ong, and Chen~Change Loy.
\newblock Unsupervised object-level representation learning from scene images.
\newblock {\em arXiv}, 2021.

\bibitem{xu2021end}
Mengde Xu, Zheng Zhang, Han Hu, Jianfeng Wang, Lijuan Wang, Fangyun Wei, Xiang
  Bai, and Zicheng Liu.
\newblock End-to-end semi-supervised object detection with soft teacher.
\newblock In {\em Proceedings of the IEEE/CVF International Conference on
  Computer Vision}, pages 3060--3069, 2021.

\bibitem{zhang2022dino}
Hao Zhang, Feng Li, Shilong Liu, Lei Zhang, Hang Su, Jun Zhu, Lionel~M Ni, and
  Heung-Yeung Shum.
\newblock Dino: Detr with improved denoising anchor boxes for end-to-end object
  detection.
\newblock {\em arXiv preprint arXiv:2203.03605}, 2022.

\bibitem{zhao2021self}
Yucheng Zhao, Guangting Wang, Chong Luo, Wenjun Zeng, and Zheng-Jun Zha.
\newblock Self-supervised visual representations learning by contrastive mask
  prediction.
\newblock In {\em ICCV}, pages 10160--10169, 2021.

\bibitem{zheng2019re}
Meng Zheng, Srikrishna Karanam, Ziyan Wu, and Richard~J Radke.
\newblock Re-identification with consistent attentive {S}iamese networks.
\newblock In {\em CVPR}, pages 5735--5744, 2019.

\bibitem{zhu2020deformable}
Xizhou Zhu, Weijie Su, Lewei Lu, Bin Li, Xiaogang Wang, and Jifeng Dai.
\newblock Deformable detr: Deformable transformers for end-to-end object
  detection.
\newblock {\em ICLR}, 2020.

\bibitem{zitnick2014edge}
C~Lawrence Zitnick and Piotr Doll{\'a}r.
\newblock Edge boxes: Locating object proposals from edges.
\newblock In {\em ECCV}, pages 391--405. Springer, 2014.

\end{thebibliography}
